\pdfoutput=1
\documentclass[11pt]{article}

\usepackage[final]{acl}

\usepackage{times}
\usepackage{latexsym}

\usepackage[T1]{fontenc}
\usepackage[utf8]{inputenc}

\usepackage{microtype}

\usepackage{inconsolata}

\usepackage{microtype}
\usepackage{textcase}

\usepackage[T1]{fontenc}
\usepackage[utf8]{inputenc}
\usepackage{caption}

\usepackage{tabularx} % in the preamble
\usepackage{amssymb}% need for xmark & xmark
\usepackage{amsmath}
\usepackage{mathtools}
\usepackage{pifont}% http://ctan.org/pkg/pifont

\usepackage{ragged2e}
\usepackage{booktabs}

\usepackage{xcolor}
\usepackage{colortbl}
\newcommand{\drop}[1]{\textcolor{gray}{\textsubscript{--#1}}}
\newcommand{\rise}[1]{\textcolor{gray}{\textsubscript{+#1}}}

\makeatletter
\newcommand{\removelatexerror}{\let\@latex@error\@gobble}
\makeatother

\usepackage{color}
\definecolor{amber}{rgb}{1.0, 0.75, 0.0}
\definecolor{ao(english)}{rgb}{0.0, 0.5, 0.0}
\definecolor{amaranth}{rgb}{0.9, 0.17, 0.31}
\newcount\Comments  % 0 suppresses notes to selves in text
\newcommand{\kibitz}[2]{\ifnum\Comments=1\textcolor{#1}{#2}\fi}

\newcommand{\haozhu}[1]{}

\newcommand{\vsiont}[1]{}

\usepackage{subcaption}

\newcommand{\method}{\texttt{LaRS}\space}

\newcommand{\rmethod}{\method}
\newcommand{\qmethod}{Retrieval-Q}

\def\RA{R} % rationale
\def\Qt{Q_{\text{test}}}

\usepackage{enumerate}

\usepackage{fancyhdr}
\usepackage{color}
\usepackage{url}
\usepackage{multirow}
\usepackage{graphicx} %Loading the package
\usepackage{times}
\usepackage{latexsym}
\usepackage{multicol}
\usepackage{multirow}
\usepackage{algorithmic}
\usepackage{algorithm}% http://ctan.org/pkg/algorithms
\usepackage{placeins}

\usepackage{lipsum}
\usepackage{titlesec}
\titlespacing\section{0pt}{5pt plus 0pt minus 1pt}{1pt plus 1pt minus 1pt}
\titlespacing\subsection{0pt}{2pt plus 0pt minus 1pt}{0pt plus 0pt minus 0pt}
\titlespacing\subsubsection{0pt}{2pt plus 0pt minus 1pt}{0pt plus 0pt minus 0pt}
\titlespacing{\paragraph}{0pt}{1pt}{0pt}[0pt]  

\usepackage{etoolbox}
\makeatletter
\preto{\@tabular}{\parskip=5pt}
\makeatother

\allowdisplaybreaks

\usepackage{enumitem}
\setlist[itemize]{leftmargin=*}

\usepackage{caption}
\usepackage{url}

\newcommand{\fref}[1]{Fig.~\ref{#1}}

\newcommand{\sref}[1]{Section~\ref{#1}}

\newcommand{\cref}[1]{Condition~(\ref{#1})}

\usepackage{amsfonts,amssymb,bm}
\usepackage{pifont} % http://ctan.org/pkg/pifont
\usepackage{graphicx,epsfig,fancybox} % figures & graph
\usepackage{float}
\usepackage{color} % color
\usepackage{multirow}

\newcommand{\revise}[1]{\textcolor{blue}{}}
\newcommand{\revised}[1]{\textcolor{blue}{}}

\usepackage{adjustbox}
\usepackage{array}
\usepackage{booktabs}

\DeclareMathAlphabet{\mathsfit}{\encodingdefault}{\sfdefault}{m}{sl}
\SetMathAlphabet{\mathsfit}{bold}{\encodingdefault}{\sfdefault}{bx}{n}

\usepackage{algorithm}
\usepackage{algorithmic}
\usepackage{amsmath}               
  {
      \newtheorem{assumption}{Assumption}
  }
\usepackage{amsmath}
\usepackage{multirow}
\usepackage{booktabs}

\usepackage{amsmath}               
  {
      \newtheorem{definition}{Definition}
  }
              
  {
      \newtheorem{theorem}{Theorem}
  }
  
  {
      
  }

\usepackage[bb=dsserif]{mathalpha}
\usepackage{bm}

\DeclareMathOperator*{\argmax}{arg\,max}

\title{LaRS: Latent Reasoning Skills for Chain-of-Thought Reasoning}

\author{{Zifan Xu\textsuperscript{1}\thanks{Majority of this work was done when Zifan Xu was an intern at Amazon Web Service during the summer 2023. A portion of this work has taken place in the Learning Agents Research
Group (LARG) at UT Austin.  LARG research is supported in part by NSF
(FAIN-2019844, NRT-2125858), ONR (N00014-18-2243), ARO
(W911NF-23-2-0004, W911NF-17-2-0181), Lockheed Martin, and UT Austin's
Good Systems grand challenge.  Peter Stone serves as the Executive
Director of Sony AI America and receives financial compensation for
this work.  The terms of this arrangement have been reviewed and
approved by the University of Texas at Austin in accordance with its
policy on objectivity in research.}, Haozhu Wang\textsuperscript{2}, Dmitriy Bespalov\textsuperscript{2}, Xian Wu\textsuperscript{2}, Peter Stone\textsuperscript{1,3}, Yanjun Qi\textsuperscript{2}} \\ \\
{\textsuperscript{1}The University of Texas at Austin}, \textsuperscript{2}Amazon Web Service, \textsuperscript{3}Sony AI }

\begin{document}
\maketitle

\begin{abstract}
Chain-of-thought (CoT) prompting is a popular in-context learning (ICL) approach for large language models (LLMs), especially when tackling complex reasoning tasks. Traditional ICL approaches construct prompts using examples that contain questions similar to the input question. However, CoT prompting, which includes crucial intermediate reasoning steps (rationales) within its examples, necessitates selecting examples based on these rationales rather than the questions themselves. Existing methods require human experts or pre-trained LLMs to describe the skill, a high-level abstraction of rationales, to guide the selection. These methods, however, are often costly and difficult to scale. Instead, this paper introduces a new approach named \textbf{La}tent  \textbf{R}easoning \textbf{S}kills  (\texttt{LaRS}) that employs unsupervised learning to create a latent space representation of rationales, with a latent variable called a \emph{reasoning skill}. Concurrently, \method learns a \emph{reasoning policy} to determine the required reasoning skill for a given question. Then the ICL examples are selected by aligning the reasoning skills between past examples and the question. This approach is theoretically grounded and compute-efficient, eliminating the need for auxiliary LLM inference or manual prompt design. Empirical results demonstrate that LaRS consistently outperforms SOTA skill-based selection methods, processing example banks four times faster, reducing LLM inferences during the selection stage by half, and showing greater robustness to sub-optimal example banks. Our code is publicly available \href{https://github.com/Daffan/lars}{here}. %saving thousands of LLM inferences and significantly reducing the time required to process the example bank.

\end{abstract}

\section{Introduction}
\label{sec::intro}
Large Language Models (LLMs) exhibit remarkable capabilities in solving various downstream tasks through in-context learning (ICL)~\cite{Brown2020LanguageMA}, even without being explicitly trained on the distribution of in-context examples~\cite{Vaswani2017AttentionIA,Devlin2019BERTPO, Rae2021ScalingLM,Chowdhery2022PaLMSL, Wei2022EmergentAO}. Using in-context learning, LLMs generate output for an input query by conditioning on a prompt that contains a few input-output \emph{demonstrations}. %\footnote{A demonstration in ICL prompting refers to an input-output pair whose input is a question to LLM, and output is the response by LLM. In CoT prompting, input is a question, output is rationale, so each "demonstration" in CoT is a question-rationale pair.}.

\begin{figure}[t]
    \centering
    \begin{subfigure}[t]{0.9\columnwidth}
        \centering
        \includegraphics[width=\columnwidth]{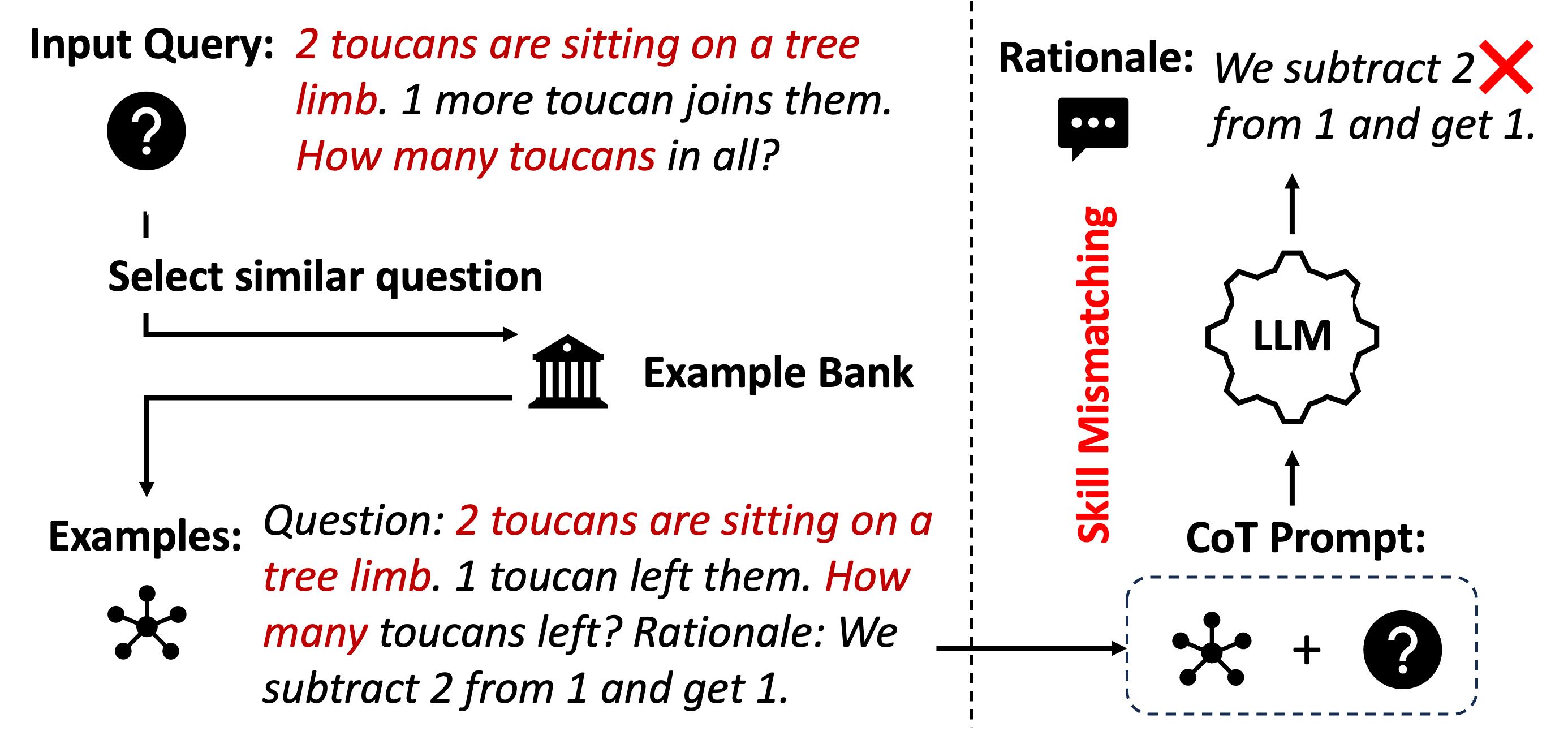}
        \caption{Question-similarity-based selection.}
        \label{fig:similarity_example}
    \end{subfigure}%
    \newline
    \begin{subfigure}[t]{0.9\columnwidth}
        \centering
        \includegraphics[width=\columnwidth]{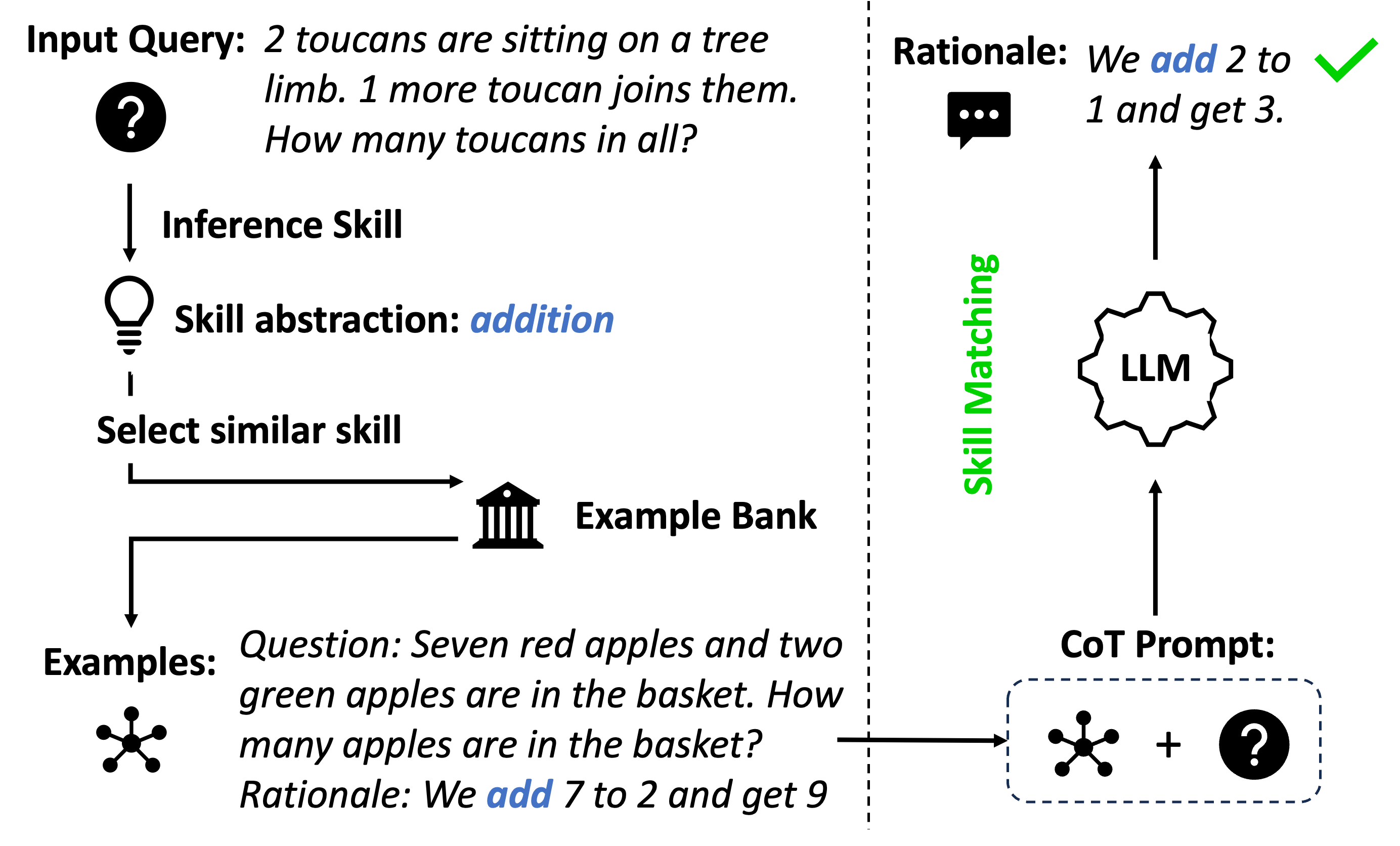}
        \caption{Skill-based selection.}
        \label{fig:skill_example}
    \end{subfigure}
    \caption{CoT prompting with examples selected by (a) similar questions and (b) similar skills that (mis)match the skills in their rationales.}
    \label{fig:similarity_vs_skill}
\end{figure}

\begin{figure*}[htb!]
     \centering
     \includegraphics[width=0.90\textwidth]{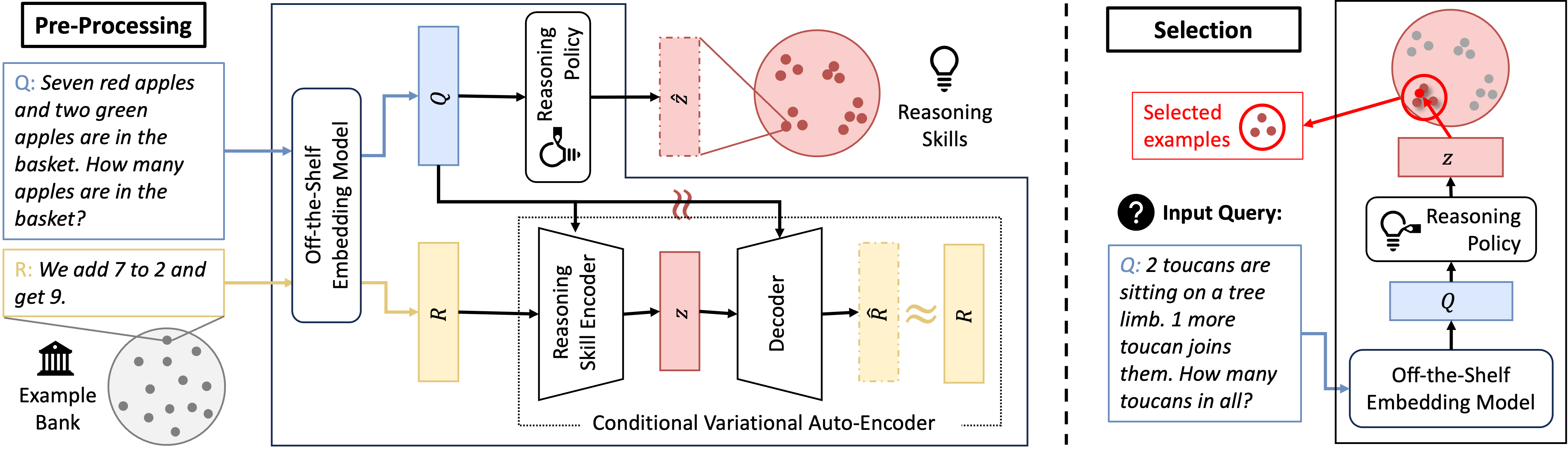}
     \caption{An overview of \method including a pre-processing stage (left) and a selection stage (right).\vspace{-14pt}}
     \label{fig::rsd_overview}
\end{figure*}
Reasoning tasks have proven to be particularly difficult for language models and NLP  in general~\cite{Rae2021ScalingLM,Bommasani2021OnTO,Nye2021ShowYW}. In the recent literature, chain-of-thought (CoT) prompting, an ICL method, has been proposed to improve LLMs on a wide spectrum of reasoning tasks by guiding LLMs to produce a sequence of intermediate steps (rationale) for generating a (better) final answer~\cite{Cobbe2021TrainingVT,Wei2022ChainOT,Suzgun2022ChallengingBT}. 
The prompts for CoT are composed of \emph{demonstrations} that contain not only input and output, but also the rationales for why the output holds.

The core challenge for ICL lies in designing effective demonstrations to prompt LLMs. Much evidence has indicated the significant impact of demonstrations on the performance of ICL~\cite{Lu2021FantasticallyOP, Liu2021WhatMG}. 
To form a prompt, one important setting considers selecting demonstrations from an existing example bank, termed demonstration selection~\cite{dong2022survey}. 
While a variety of methods exist in the ICL literature for automating this process, CoT prompts are distinct in that they include not only questions and answers but also specially-designed rationales. \emph{This distinction highlights the importance of rationales in selecting demonstrations for CoT prompting.} Specifically, CoT prompting should select demonstrations that illustrate relevant skills within their rationales to effectively address a given question.
For instance, in solving math word problems (as depicted in Fig.~\ref{fig:similarity_vs_skill}), a useful rationale involves computing addition to get the correct answer. Selecting few-shot examples based on the question similarity (Fig.~\ref{fig:similarity_example}) might lead to examples showcasing subtraction and generate incorrect rationales. However, skill-based selection (Fig.~\ref{fig:skill_example}) can align the skills between examples and the given question, which leads to correct answers guided by relevant rationales.
%Thus, prompting LLMs with demonstrations that feature addition is likely to yield better results for problems necessitating this skill, compared to those requiring subtraction.

To achieve such a skill-based demonstration selection, \citet{an2023skill} introduces \texttt{Skill-KNN}, which employs pre-trained LLMs to generate skill descriptions. Then, the few-shot examples are selected based on the embedding of the skill descriptions computed by another pre-trained embedding model. Although this approach is straightforward, the LLM-generated skill descriptions can be somewhat arbitrary, heavily relying on the manually crafted prompts. This reliance constrains its wider applicability across diverse reasoning tasks. Moreover, the approach requires to generate a unique skill description for each example, which limits its scalability to larger example banks.

Rather than relying on LLMs, we introduce \textbf{La}tent \textbf{R}easoning \textbf{S}kill Discovery (\texttt{LaRS}), a new skill-based demonstration selection method. This approach learns skills as latent space representations of rationales through unsupervised learning. The essence of \method lies in a unique formulation for the generation of rationales, which we term the latent \emph{skill model}. This model, inspired by the principles of topic models~\cite{Xie2021AnEO}, conditions the generation of a rationale on both a given question and a latent variable, called a \emph{reasoning skill}. This latent variable embodies a high-level abstraction of the rationales, such as formats, equations, or knowledge.
\begin{figure}[thb!]
    \centering
    \includegraphics[width=\columnwidth]{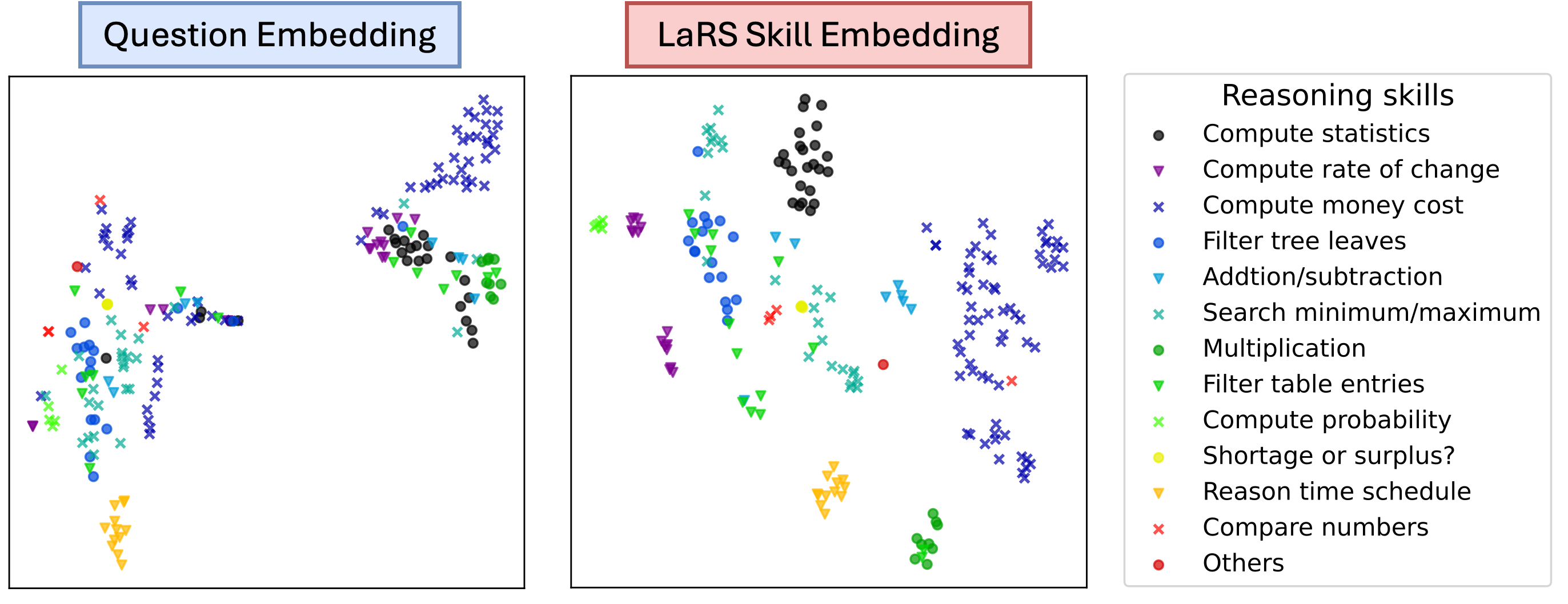}
    \caption{t-SNE projections of question embedding and LaRS reasoning skill embedding of the exmaples from TabMWP \cite{Lu2022DynamicPL} dataset. The 12 different colors correspond to 12 skill labels annotated by human.}
    \label{fig:clustering_small}
\end{figure}
\vspace{-10pt}

Under the skill model formulation, \method utilizes a Conditional Variational Auto-encoder (CVAE) to approximate the generation of rationales on a small dataset from the example bank. As a result, two probabilistic models can be learned concurrently: (1) a \emph{reasoning skill encoder} that maps an example to the actual reasoning skills demonstrated in the rationale; and (2) a \emph{reasoning policy} that predicts the reasoning skills required for a particular question. This method of learning through a CVAE, especially when applied to a small dataset from the example bank, is both cost-efficient and fast compared to \texttt{Skill-KNN}. Fig.~\ref{fig::rsd_overview} presents an overview of \texttt{LaRS}. In addition, Figure \ref{fig:clustering_small} shows the learned reasoning skill embedding (right) that effectively separates examples with different skill labels, while the off-the-shelf question embedding does not.

The efficacy of \method is evaluated on four different benchmarks based on five backbone LLMs with varying scales. The method is also compared with baseline approaches, including an oracle method that assumes access to ground truth rationales. \method consistently outperforms \texttt{Skill-KNN} and also matches the oracle performance in almost half of the experiments. In addition, \method reduces half of the LLM inference, eliminates the need of human prompt design, and maintains better robustness to sub-optimal example banks. A summary of this paper's contribution is as follows: 
\begin{itemize}[noitemsep,topsep=0pt]
    \item We propose \texttt{LaRS}, a novel unsupervised demonstration selection approach for CoT prompting, and empirically verify its effectiveness through large scale experiments.
    \item We introduce the latent skill model, a plausible formulation for CoT reasoning, which has illuminated a deeper understanding of CoT prompting.
    \item We present theoretical analyses of the optimality of the latent-skill-based selection method.
\end{itemize}

\begin{figure}[t]
    \centering
    \includegraphics[width=\columnwidth]{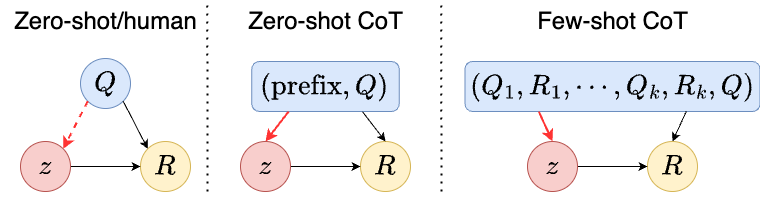}
    \caption{Causal graphs for prompting with  zero-shot/human (left), zero-shot CoT (middle), and few-shot CoT (right) for generating rationales via skills. The dashed arrow from $Q$ to $z$ indicates possible sub-optimal inference of the reasoning skills from both human and zero-shot LLM generations.}
    \label{fig:casualg}
\end{figure}

\section{Related Work}
%This section discusses related work in two different directions.
\subsection{CoT Reasoning}
CoT prompting is a special prompt design technique that encourages LLMs to generate intermediate rationales that guide them towards providing accurate final answers. These rationales can exhibit remarkable flexibility in their styles. For instance, the original work by \cite{Wei2022ChainOT} specially designs rationales in the in-context demonstrations to suit different reasoning tasks. Moreover, novel prompt designs that highlight diverse formats of the rationales have emerged to enhance CoT prompting. For example, \cite{kojima2022large} proposed Program of Thoughts (PoT) that disentangles textual reasoning from computation, with the latter specially handled through program generation.

In contrast to manual design, our method \method can be thought of as automatic discovery of diverse rationale styles from an example bank. This method can also dynamically select reasoning skills based on the specific questions.
Worth noting, \cite{chen2023skills} introduces SKills-in-Context (SKiC), which confines rationale generation to predefined ``skills'' within the prompt. Although sharing a similar motivation to \texttt{LaRS}, we emphasize two crucial distinctions: (1) while SKiC relies on manual ``skills'' design, \method automatically discovers them, (2) SKiC presents a full list of ``skills'' in the prompt, allowing LLMs to select from them, whereas \method learns the skill selection from the example bank, explicitly instructing LLMs on which skill to employ through in-context examples.

\subsection{Demonstration Selection}
Demonstration selection refers to a special setting, where the prompts are constructed by selecting examples from an example bank. In this context, our \method aligns with the paradigm of unsupervised demonstration selection, which involves designing heuristics for this selection process. A variety of heuristics have been explored, including similarity~\cite{Gao2021MakingPL,Hu2022InContextLF}, diversity~\cite{Zhang2022AutomaticCO}, coverage~\cite{Gupta2023CoveragebasedES}, and uncertainty~\cite{Diao2023ActivePW}. Among these, \texttt{Skill-KNN} (\cite{an2023skill}) shares the closest resemblance to our approach. However,  \texttt{Skill-KNN} relies on pre-trained LLMs to provide ``skill'' annotations, which could be arbitrary and resource-intensive, requiring extensive inferences of LLMs and human prompt design. In contrast, \method automatically discovers reasoning skills by learning a lightweight CVAE represented by two-layer MLPs and standard loss function. In addition, the selections based on these discovered reasoning skills are theoretically-grounded based on the latent skill model and the theoretical analyses presented in this paper. 
%\zifan{Removing supervised selection to save the space...}
%Another paradigm is supervised demonstration selection. Research in this direction treats demonstration selection as a black-box problem, and optimizes the selection with "training signals", such as the answer correctness~\cite{Lu2022DynamicPL, Rubin2021LearningTR, li2023unified, zhang2022active}. Nevertheless, these "training signals" could be expensive to collect and are sometimes inaccessible.

%Discuss supervised and unsupervised selection respectively... Unsupervised selection is universal and applicable to CoT prompting, but is costly \zifan{highlight \cite{Li2023UnifiedDR}, SOTA supervised selection}. While for unsupervised selection, we argue existing heuristic might be directly applicable to CoT reasoning. \zifan{highlight \cite{an2023skill}, it shares similar motivation, but require manual design and LLM-inference}.

%\subsection{Latent space theory for in-context learning}
%\zifan{Discuss the difference between \cite{Wang2023LargeLM}.}

\section{Formulation}
In this section, we formally describe the \emph{skill model}, a new formulation for explaining the generation of rationales in CoT reasoning. In \sref{sec:skill_model}, the skill model is first introduced to describe the human-generated rationales. Then, \sref{sec:cot_prompting} illustrates how the skill model can be adapted to LLM-generated rationales. Finally, leveraging the concept of reasoning skill as outlined in the skill model, a new latent-skill-based demonstration selection method is formally described in \sref{sec:skill-based}.

\subsection{Skill Model}
\label{sec:skill_model}
Let $\mathcal{X}$ be the set of all sequences of tokens, $\mathcal{Z}$ be the continuous vector space of latent reasoning skills, and $P_H$ denotes the probability distribution of real-world natural language. CoT reasoning is to generate a rationale $\RA \in \mathcal{X}$ given a question $Q \in \mathcal{X}$, whose correctness\footnote{For math word problems, whose answers are discrete labels, the correct rationale should contain the correct answer label as the final step. For code generation, the correct rationale should be the correct code.} can be verified by an indicator function $\mathbb{1}(\RA, Q) := \mathbb{1}(\RA \text{ is the correct rationale for } Q)$. 

The skill model assumes that the real-world conditional distribution of $\RA$ given $Q$ can be described as follows:
\begin{equation}
\label{eq:skill_model}
\scalebox{0.95}{%
$P_H(\RA\mid Q) = \int_{\mathcal{Z}} P_H(\RA\mid z, Q)P_H(z\mid Q) dz $
}
\end{equation}
where, $P_H(z\mid Q)$ is the posterior of selecting latent reasoning skills in human reasoning, called a reasoning policy. $P_H(\RA \mid z, Q)$ is the posterior distribution of generating $\RA$ given a question $Q$ and a reasoning skill $z$. A causal graph illustrating such a generation process involving a latent reasoning skill $z$ is presented in Fig. \ref{fig:casualg} on the left.
%We begin by describing the skill model, a plausible formulation for explaining the generation of reasoning skills adopted from a modified topic model~\cite{wang2023large}. Consider $\RA \in \mathcal{X}$ representing a rationale, $\Qt \in \mathcal{X}$ representing a test question, $z \in \mathcal{Z}$  representing a latent variable of reasoning skill, with $\mathcal{X}$ being the space of all possible token sequences and $\mathcal{Z}$ being a potentially high-dimensional continuous latent space. Then, the generation of $S$ can be formulated as a probability distribution conditioned on both $\Qt$ and $z$ that can be inferred from $\Qt$:

%Here, $\pi_H(z|Q):\mathcal{X} \mapsto \Delta(\mathcal{Z})$ represents a reasoning policy that maps a test question to a probability distribution of reasoning skills. The subscript $H$ underscores the human-generated probability.

Unlike \cite{wang2023large}, this formulation considers a dependency of $z$ on $Q$ reflecting a preference for selecting particular reasoning skills to solve a given question. We justify this formulation as follows:
\begin{enumerate}[noitemsep,topsep=0pt]
    \item Rationales can exhibit remarkable flexibility, manifesting diverse formats, topics, and knowledge, which can naturally be abstracted into the high-level concepts of reasoning skills.
    \item The selection of these skills is not bound by strict determinism. For instance, diverse reasoning paths and formats could all contribute toward finding the correct final answer. Therefore, real-world data is a mixture of diverse skills captured by a stochastic reasoning policy $P_H(z\mid Q)$.
\end{enumerate}
%Fig. \ref{fig:cot_reasoning} (left) illustrates such a generation process for a simple math word problem. 

\subsection{CoT prompting}
\label{sec:cot_prompting}
LLMs are pre-trained conditional generators. Given an input query $X \in \mathcal{X}$, the conditional distribution of an output $Y \in \mathcal{X}$ generated by LLMs can be written as $P_M(Y \mid X)$. LLMs are usually trained on generic real-world data distribution such that $P_M(Y \mid X) \approx P_H(Y \mid X)$.

Prior studies have presented an implicit topic model formulation in explaining the in-context learning mechanisms of LLMs~\cite{wang2023large,Xie2021AnEO}. %Specifically, instead of directly generating output based on input, this formulation suggests an implicit inference of ``topics'', based on which, the output is generated combined with the input. 
Similarly, we posit that LLMs can be viewed as implicit skill models for generating rationales. To elaborate, when generating rationales, LLMs' conditional distribution $P_M(\RA \mid Q)$ can be extended as follows (with illustrations in \fref{fig:casualg} on the left):
\begin{align}
\scalebox{0.95}{%
    $P_M(\RA\mid Q) = \int_{\mathcal{Z}} P_M(\RA \mid z, Q) P_M(z\mid Q)dz$
}
\end{align}
This implicit skill model assumes that LLMs also infer reasoning skills $z$, which resembles the  real-world generation of rationales.
%Under the implicit skill model formulation, the generation by LLMs also entails the inference of a reasoning skill $z$. The probability of generating $S$ is conditioned on both $z$ and $\Qt$.
%\zifan{add some explanation here}
%\begin{assumption}
%\label{assume:M2H}
%LLMs learn the exact real-world conditional distribution, e.g., $P_M(\RA\mid z, Q) = P_H(\RA\mid z, Q)$ and $P_M(z\mid Q) = P_H(z\mid Q)$.
%\end{assumption}

The above formulation only encompasses the zero-shot generation of rationales. In practice, prompts are commonly provided to guide LLMs' generation. In general, two CoT prompting strategies exist: zero-shot CoT, employing a prompt comprising a short prefix and a test question, and few-shot CoT, employing a prompt containing pairs of questions and rationales. Denoting $pt \in \mathcal{X}$ as a prompt, a unified formulation for both prompting strategies can be derived as follows:
\begin{equation}
\scalebox{0.95}{%
    $P_M(\RA\mid pt) = \int_{\mathcal{Z}} P_M(\RA\mid z,Q) P_M(z\mid pt)dz$ %\\
} 
    %&\text{0-shot CoT: } pt = (\text{prefix}, Q) \text{ or } (Q, \text{prefix}) \nonumber \\
    %&\text{$k$-shot CoT: } pt = (Q_1, \RA_1, \cdots, Q_k, \RA_k, Q) \nonumber
    \label{eq:cot_promting}
    \vspace{-5pt}
\end{equation}
\text{\qquad 0-shot CoT: } $pt = (\text{prefix}, Q) \text{ or } (Q, \text{prefix})$ \newline
\text{\qquad $k$-shot CoT: } $pt = (Q_1, \RA_1, \cdots, Q_k, \RA_k, Q)$

Here, the formulation is simplified such that the use of prompts only influences the probability distribution of $z$. For instance, a prefix specifying the generation's format can be interpreted as specifying the reasoning skill $z$ by shaping the distribution from $P_M(z\mid Q)$ to $P_M(z\mid pt)$. This simplification aligns with empirical evidence suggesting that in-context examples serve as mere pointers to retrieve already-learned knowledge within LLMs~\cite{Shin2020ElicitingKF,min2022rethinking,wang2022towards}.

Drawing upon this formulation, we can gain insight into the failure of zero-shot generation. \emph{In general, real-world data is inherently noisy, indicating that the reasoning policy $P_H(z\mid Q)$ may be sub-optimal, and the reasoning skills are not chosen to maximize the accuracy of answering a test question.} Trained on this generic real-world data distribution, $P_M(z\mid Q)$ could also be sub-optimal, leading to the failure of zero-shot generation. On the other hand, CoT prompting improves the reasoning performance by shaping the distribution of reasoning skills using carefully-designed prompts that contain either prefix or few-shot examples.

% Here, we assume that LLM's generation of $S$ also involves inferring a latent variable of reasoning skill $z$, and the probability of generating $S$ is only conditioned on $z$ and $\Qt$. Such a formulation limits that the choices of demonstrations in the prompt only influences the preference of reasoning skills, while the generation of $S$ is independent to the choice of demonstrations.

\subsection{Skill-Based Demonstration Selection}
\label{sec:skill-based}
The analysis above suggests that the key to the success of CoT prompting is to design an effective prompt that improve upon the posterior distribution of human's preference of reasoning skills $P_H(z \mid Q)$. To design an effective prompt, the demonstration selection problem assumes access to an example bank of question-rationale pairs, denoted as $\mathcal{D}_E = \{(\RA, Q)\}$. This example bank is usually specially-crafted and has a distribution different from the real-world distribution. Denoting $P_E$ as the distribution of the example bank, $\RA$ is distributed according to $P_E(\RA \mid Q)$ for all $(\RA, Q) \in \mathcal{D}_E$.

%We start by defining an expert distribution in contrary to this sub-optimal real-world distribution as follows:
%\begin{definition} Expert distribution $P^*$ is given by
%\begin{align}
%    P^* = \argmax_{P} \int_{\mathcal{Z}} \mathbb{1}(\RA, Q) P(\RA \mid z, Q) P(z \mid Q) dz
%\end{align}
%\end{definition}

Given $\mathcal{D}_E$, the demonstration selection is to select a few question-rationale pairs from $\mathcal{D}_E$. Assuming that each selected demonstration is i.i.d, a demonstration selection method can be uniquely defined as a probabilistic model $g(Q, \RA|\Qt):= \mathcal{X}  \mapsto \Delta(\mathcal{X})$ that maps a test question $\Qt$ to a probability distribution of demonstrations. Then, we can formally define the skill-based demonstration selection method as follows:
\begin{definition} Skill-based demonstration selection is given by
\begin{align}
\scalebox{0.9}{%
$g_{\text{skill}}(Q, \RA\mid Q_{\text{test}}) = \int_{\mathcal{Z}} P_E(Q, \RA\mid z) P_E(z\mid Q_{\text{test}})dz \nonumber$
}
\end{align}
\label{def:retrieval_RSD}
\end{definition}
\vspace{-16pt}
Intuitively, this selection method maximizes the probability of a selected demonstration showcasing the reasoning skill that is likely to be chosen according to $P_E(z\mid Q)$. Since the example bank is usually specially-crafted and contains rationales showcasing ``better'' reasoning skills, the in-context examples that align with $P_E(z \mid Q)$ are intuitively more effective. In Section \ref{sec:theory}, we provide theoretical analysis of the optimality of this skill-based selection when conditioned on certain ideal assumptions of the example bank and LLMs.

%Consider a few-shot demonstration selection for CoT prompting given an example bank $\mathcal{D}_E = \{(\RA, Q)\}$. We assumes that the examples are generated by experts, meaning that $\RA \sim P^*(\RA \mid Q), \forall (\RA, Q) \in \mathcal{D}_E$. Although idealistic, we believe that this setting aligns with real-world demonstration selection problems, where specially-crafted demonstrations are usually provided as the example bank.

% \input{content/01-figure-tsne}

\section{Method}
%\begin{figure*}[hbt!]
%    \centering
%    \includegraphics[width=\textwidth]{content/figures/fig3_v3.pdf}
%    \caption{An overview of \method and the demonstration selection process. \qnote{this may go appendix}}
%    \label{fig:RSD}
%\end{figure*}
%Based on the theoretical analyses presented in the previous section, we can conclude that with an expert reasoning policy and a reasoning skill encoder, we can effectively select demonstrations from an expert demonstration dataset. 
To enable the skill-based demonstration selection (Definition \ref{def:retrieval_RSD}), we introduce our approach \method, which involves learning  a conditional variational autoencoder (CVAE) to approximate $P_E$ using the data from the example bank $\mathcal{D}_E$. We then outline a practical demonstration selection process aligning with the skill-based selection. The schematic overview of \method (right) and the corresponding demonstration selection process (left) are illustrated in Figure \ref{fig::rsd_overview}.

\subsection{Latent Reasoning Skill Discovery}
\label{sec:RSD}

%\jane{ADDING:
%The variational autoencoder (VAE) has been a popular approach for unsupervised representation learning. The VAE include two coupled, but independently parameterized models: the encoder  model, and the decoder model. 
%The encoder model delivers to the decoder model an approximation of its posterior over latent variables. 
%The decoder model of VAE is a Bayesian network $p(\bx|\bz) p(\bz)$ generating data $\bx$ from latent $\bz$. Similarly, the encoder model of VAE is also a conditional Bayesian network of the form $q(\bz|\bx)$, mapping $\bx$ to $\bz$. VAE uses a different deep neural network to model each of the two conditionals, e.g. $\bz|\bx \sim g(\bx,\bbeps)$, with $g$ a neural network module and $\bbeps$ a noise random variable. In this paper, we train VAE using the classical variational expectation maximization and using the reparameterization trick. More specifically, we use the CVAE that extends the standard VAE framework by conditioning on auxiliary variables, e.g., in our case, ????.}

The conditional variational autoencoder (CVAE) has emerged as a popular approach for modeling probabilistic conditional generation. As one specific case, the skill model, introduced in this paper, can effectively be represented as a CVAE. Therefore, we introduce \method that employs a CVAE to approximate the generation of rationales using the data from the example bank $\mathcal{D}_E = \{(Q, \RA)\}$.

In particular, this CVAE includes three coupled models: an encoder model, a decoder model, and a reasoning policy model, independently parameterized by $\omega$, $\psi$, and $\phi$ respectively. Drawing from the notations introduced in the skill model, the reasoning policy model is a conditional Bayesian network $\pi_\phi(z\mid Q)$, determining the posterior distribution of latent reasoning skill $z$ given a question $Q$. The decoder model is  also a conditional Bayesian network $p_\psi(\RA \mid z, Q)$ that generates a rationale $\RA$, conditioned on both $Q$ and $z$, where $z$ is sampled from $\pi_{\phi}(z\mid Q)$. Finally, the encoder model $q_\omega(z\mid Q, \RA)$ is another conditional Bayesian network, mapping a question-rationale pair to $z$. In this paper, we train this CVAE using classical variational expectation maximization and the reparameterization trick.

%According to the skill model (Equation \ref{eq:skill_model}), the probability $P_E(\RA|z, Q)$ of generating a rationale $\RA$ is conditioned on a question $Q$ and a latent variable of reasoning skill $z$, where a reasoning policy $\pi_E(z|Q)$ defines a prior distribution of $z$. Notice that such a probabilistic generation process matches exactly the conditional generative model (CGM) proposed by \cite{sohn2015learning}. 

%CGM can be effective approximated by a conditional variational autoencoder, which involves an additional recognition network to approximate the posterior distribution. Therefore, we employ CVAE to approximate the skill model, and introduce the three learnable components in this CVAE as follows: 
%This generation process can be effectively represented by a CVAE that contains the following three learnable components:
%\paragraph{Reasoning Policy (Prior Predictor):} $\pi_{\phi}(z|Q)$ encodes the question into a distribution of reasoning skills and gives out the parameters of that distribution.
%\paragraph{Encoder (Posterior Predictor):} $P_{\omega}(z|Q, \RA)$ encodes the concatenation of a question and a rationale into a distribution of reasoning skills and gives out the parameters of that distribution.
%\paragraph{Decoder (Text Generator): } $P_\psi(\RA|z, Q)$ is the skill-conditioned text generator that generates the rationale from a given reasoning skill $z$ and a question $Q$.

Specifically, the classical variational expectation maximization optimizes a loss function as follows:
\scalebox{0.95}{\parbox{1.04\columnwidth}{%
\begin{align}
\label{eq:loss}
    %\max_{\phi, \omega, \psi} \mathbb{E}_{(Q, \RA) \sim \mathcal{D}, z \sim P_{\omega}(z|Q,\RA)}[\log P_\psi(\RA |z, Q)]  \\
    %- \mathbb{E}_{(Q,\RA) \sim \mathcal{D}}[\text{D}_{\text{KL}}(P_{\omega}(z|Q,\RA)\parallel \pi_\phi(z|Q))] \leq \epsilon_{\text{KL}} \nonumber
    &\mathcal{L}_{\text{CVAE}}(\phi, \omega, \psi) = \mathcal{L}_{\text{recon}} + \mathcal{L}_{\text{KL}}\\
    &\mathcal{L}_{\text{recon}} = -\mathbb{E}_{(Q, \RA) \sim \mathcal{D}_E, z \sim q_{\omega}(\mid Q,\RA)}[\log p_\psi(\RA |z, Q)] \nonumber \\ 
    &\mathcal{L}_{\text{KL}} = ~\mathbb{E}_{(Q,\RA) \sim \mathcal{D}_E}[\text{D}_{\text{KL}}(q_{\omega}(z\mid Q,\RA)\parallel \pi_\phi(z\mid Q))] \nonumber
\end{align}
}}
%Intuitively, \method learns a decoder $P_{\psi}(\RA| z, Q)$ to accurately reconstruct the rationale $\RA$, conditioning on a latent variable $z$ and a question $Q$. Simultaneously, KL divergence loss (second term on the right hand side) serves as a regularization to ensure that the learned reasoning skill $z$ is predictable from the question (prior) and the question-rationale pair (posterior). This loss prevents overfitting to individual question-rationale pairs, and the learned represented becomes a high-level abstraction of rationales. Fig. \ref{fig:RSD} on the right side gives an overview of \method.
By training to minimize this loss function, $q_{\omega}$ and $\pi_\phi$ can be learned to effectively approximate the conditional distributions $P_E(z\mid Q, \RA)$ and $P_E(z\mid Q)$. It is worth noting that the decoder model acts an auxiliary model that only roughly reconstructs rationales for the purpose of training the encoder and the reasoning policy model, and is not deployed to generate rationales in the downstream tasks.

Ideally, all three models would be represented by language models, processing token sequences as input and generating token sequences as output. However, training full language models for demonstration selections can be computationally expensive. Instead, we adopt a pre-trained embedding model denoted as $f : \mathcal{X} \mapsto \Theta$, which maps the token space $\mathcal{X}$ to an embedding space $\Theta$. Consequently, the decoder model, encoder model, and reasoning policy model transform into $p_{\psi}(f(\RA)| z, f(Q))$, $q_{\omega}(z|f(Q, \RA))$, and $\pi_{\phi}(z|f(Q))$, respectively. They now condition on and generate the embeddings instead of the original tokens. In the actual implementation, we use the same feed-forward neural network to represent both $\pi_{\phi}$ and $q_{\omega}$, predicting the mean and variance of Gaussian distributions of latent reasoning skills. On the other hand, $p_{\psi}$ is a feed-forward neural network that deterministically predicts a value in the embedding space.

\subsection{Demonstration Selection}
%According to Theorem \ref{theorem:optimal}, the demonstration selection problem can be optimized by a reasoning-skill-based selection strategy given by Definition \ref{def:retrieval_RSD}. 
Since the distribution $P_E(Q, \RA \mid z)$ in Definition \ref{def:retrieval_RSD} is practically intractable, we propose a selection process that effectively aligns with the skill-based selection using the learned $\pi_\phi$ and $q_\omega$. For a given test question $\Qt$, the desirable reasoning skill $z_{\text{test}} = \argmax_z [\pi_{\phi}(z|f(Q_{\text{test}}))]$ can be computed using the reasoning policy. Subsequently, each example from the example bank can be scored based on the cosine similarity between $z_{\text{test}}$ and $z_{\text{post}}$, where $z_{\text{post}} = \argmax_z [q_{\omega}(z|Q,\RA))]$ represents the maximum likelihood skill of the current example. Finally, a CoT prompt can be constructed by selecting the top-$k$ examples according to the computed scores. The step-by-step procedure is outlined in Algorithm \ref{alg:retrieval_RSD}. %, with a summarized visual representation provided in Fig. \ref{fig:RSD} (left).% According to theorem \ref{theorem::IBI}, such a selection of demonstrations will maximize the accuracy of the predicted answer. \zifan{How to make sure such a learned representation is the same as the ground truth representation??}

\subsection{Theoretical Analysis}
\label{sec:theory}
In this section, we provide a theoretical analysis of the optimality of the skill-based selection by Definition \ref{def:retrieval_RSD}.

Let $P_M(\RA \mid Q, g)$ denotes LLMs' conditional distribution of a rationale $\RA$ given a test question $Q$ under a demonstration selection method $g$. $P_M(\RA \mid Q, g)$ can be extended as follows:
\scalebox{0.95}{\parbox{1.04\columnwidth}{%
\begin{align}
    &P_M(\RA \mid Q, g) \nonumber \\ 
    &= \int_{\mathcal{X}^k}P_M(\RA\mid pt)\Pi_{i=1}^k[g(Q_i, R_i\mid Q) d(Q_i, R_i)] \nonumber
\end{align}
}}
Here, each demonstrations $(Q_i, \RA_i)$ is independently sampled from $g(Q_i, \RA_i \mid Q), \forall i=1, \cdots, k$. These $k$ demonstrations form a prompt $pt = (Q_1, \RA_1, \cdots, Q_k, \RA_k, Q)$.

We want to show that $P_M(\RA \mid Q, g)$ is the optimal conditional distribution that maximizes the accuracy of rationales if the selection follows skill-based selection method or $g = g_{skill}$. We begin by defining the optimal conditional distribution as follows:
\begin{definition} Optimal conditional distribution of rationales given questions $P^*(\RA \mid Q)$ is given by:
\scalebox{0.95}{\parbox{1.04\columnwidth}{%
$$
P^*(\RA \mid Q) = \argmax_{P(\cdot \mid Q) \in \Delta(\mathcal{X})}\int_{\mathcal{X}} \mathbb{1}(\RA, Q) P(\RA \mid Q) d\RA
\label{def:optimal}
$$
}}
Here $\mathbb{1}(\RA, Q)$ is the indicator function of the correctness of $\RA$ given a question $Q$ (see \sref{sec:skill_model}).
\end{definition}
Then, we state two major assumptions as follows:
%First, we assume that all humans and LLMs rationale generators given reasoning skills and questions, meaning that the conditional distributions of generating $\RA$ given $z$ and $Q$ are the same.
\begin{assumption} Example bank is sampled from the optimal conditional distribution, or $P_E(\RA \mid Q) = P^*(\RA \mid Q)$.
\label{assump:2}
\end{assumption}
\begin{assumption} Humans and LLMs are expert rationale generators given reasoning skills and questions, meaning that \newline $P_H(\RA\mid z, Q) = P_E(\RA \mid z, Q) = P_M(\RA \mid z, Q)$.
\label{assump:1}
\end{assumption}
Assumption \ref{assump:2} is rooted in the fact that example banks are human-crafted that contains the most useful rationales for answering the questions. In Assumption \ref{assump:1}, $P_M$ capturing $P_H$ is a common assumption in the literature studying LLMs~\cite{xie2021explanation,Saunshi2020AME,Wei2021WhyDP}. $P_E(\RA \mid z, Q) = P_H(\RA \mid z, Q)$ is based on the assumption that reasoning skills are shared across humans, and the generation of rationales is identical given the same reasoning skills and questions.
%This assumption is common... \zifan{fill it.} Second, since example banks are specially-crafted that contains the most useful rationales for answering the questions, we assume that the conditional distribution $P_E(\RA\mid Q)$ maximizes the expectation of answer accuracy:
%, which can be formally stated as below:
%\begin{theorem}
%$\hat{P}_M(\RA \mid \Qt, g_{\text{\method}}) = P^*(\RA \mid \Qt)$
%\end{theorem}

Based on the above definiton and two assumptions, we prove the following theorem.
\begin{theorem} A LLM gives the optimal conditional distribution of rationales given questions:
$$P_M(\RA \mid Q, g_{skill}) = P^*(\RA \mid Q)$$
If (1) it is prompted by $k \rightarrow \infty$ in-context examples selected by the skill-based selection $g_{skill}$ defined by Definition \ref{def:retrieval_RSD}, (2) Assumption \ref{assump:1} and Assumption \ref{assump:2} hold.
\label{theorem:1}
\end{theorem}
Appendix \ref{appendix:theory} presents the proof for Theorem \ref{theorem:1}.

%First, $\hat{P}_M(\RA \mid \Qt, g_{\text{\texttt{RSD}}})$ can be extended as follows:
%\begin{align}
%    &\hat{P}_M(\RA \mid \Qt, g_{\text{\texttt{RSD}}}) \nonumber \\ 
%    &= \int_{\mathcal{X}^k}P_M(\RA\mid pt)\Pi_{i=1}^k[g_{\text{\texttt{RSD}}}(Q_i, R_i\mid \Qt) d(Q_i, R_i)] \nonumber\\
%    &= \int_{\mathcal{Z}}P_M(\RA \mid z, \Qt) P_M(z \mid \Qt) \Pi_{i=1}^k[P_{\method}(z|\Qt)] dz \nonumber
%\end{align}
%where $pt = (Q_1, R_1, \cdots, Q_k, R_k, \Qt)$ is a prompt that contains $k$ examples sampled from $g_{\text{\texttt{RSD}}}(Q, R\mid \Qt)$. In the third line, $P_{\method}(z|\Qt)$ is given by
%$$\int_{\mathcal{X}} P_M(z \mid Q, R) g_{\text{\texttt{RSD}}}(Q, R\mid \Qt) d(Q, R)$$

%\begin{assumption}
%\label{assume:star2H}
%$P^*(z \mid Q, R) = P_H(z \mid Q, R)$ and $P^*(R \mid z, R) = P_H(R \mid z, Q)$
%\end{assumption}

%\begin{proposition}
%$P_{\method}(z|\Qt) = P^*(z|\Qt)$
%\end{proposition}

\section{Experiments}
This section describes the experimental settings, baselines, metrics, and main results. 
%including benchmarks, compared selection methods, backbone models, and hyper-parameters. Lastly, the main results of these experiments are presented.
\subsection{Dataset}
For benchmarking, the selection methods are evaluated on four challenging datasets, including two datasets of Math Word Problem (MWP): \textbf{TabMWP}, \textbf{GSM8K}, one text-to-SQL dataset: \textbf{Spider}, and one semantic parsing dataset: \textbf{COGS}. 

Each dataset is split into a training set used to learn \method models and a test set used to evaluate the selection methods. While the training sets may potentially be large, we use randomly sampled 1K examples from the training set as the example bank, from which, the examples can be selected for CoT prompting. Detailed descriptions of the datasets and splitting are presented in Appendix \ref{appendix:expriments}.

To measure the performances, we use the answer accuracy for \textbf{TabMWP} and \textbf{GSM8K}, with the answers extracted by searching the texts right after a prefix \texttt{The answer is}. For \textbf{Spider}, we use the official execution-with-values accuracy\footnote{We use the official evaluation scripts for Spider in  https://github.com/taoyds/test-suite-sql-eval.}. For \textbf{COGS}, we report the exact-match accuracy for semantic parsing.

\subsection{Selection Methods}
Our method \method is compared with the following four baselines. All the hyper-parameters related to these methods are listed in Appendix \ref{appendix:expriments}.
\paragraph{\texttt{Skill-KNN}}\quad This baseline represents a state-of-the-art (SOTA) skill-based selection method. It employs pre-trained LLMs to generate skill descriptions for both the questions in the example bank and the test question. Then, the method selects examples whose skill descriptions most closely match that of the test question to form the prompt, using cosine similarity computed with a pre-trained embedding model. To examine the dependency on the LLMs' ability to generate skill descriptions, we introduce two variations: \texttt{Skill-KNN-large}, which uses the larger LLM gpt-3.5-turbo, and \texttt{Skill-KNN-small}, which uses the smaller LLM Falcon-40B-instruct. Additionally, to evaluate the effect of human-annotated skill descriptions prompting the LLMs to generate new skills, we introduce \texttt{Skill-KNN-zero}, which uses gpt-3.5-turbo to generate skill descriptions in a zero-shot fashion. \texttt{Skill-KNN-zero} closely resembles the setting of \method, as it does not rely on human prompt design. Therefore, \method is primarily compared with \texttt{Skill-KNN-zero}.
\paragraph{Random}\quad This baseline randomly selects $k$ in-context examples from the example bank. For each test question, the accuracy is reported as an average over three independent random selections.
\paragraph{Retrieval-Q}\quad This baseline employs a pre-trained embedding model to encode a test question, and selects in-context examples based on the cosine similarity between embeddings from examples' questions and the test question.
\paragraph{Retrieval-R (oracle)}\quad This baseline employs a pre-trained embedding model to encode the ground-truth rationale of a test question, and selects in-context examples based on the cosine similarity between examples' rationales and the ground-truth rationale.

\subsection{Backbones and Hyper-parameters}
In terms of the backbone models, the ICL is conducted by two OpenAI language models: gpt-4o and gpt-3.5-turbo, two Anthropic model: claude-3-sonnet and claude-3-haiku, and one smaller-scale Falcon-40B-Instruct~\cite{xu2023baize}. All the embedding is computed by a pre-trained embedding model, Deberta-v2-xlarge~\cite{he2021deberta}. We also investigate different choices of embedding model in Section \ref{sec:analysis}.

During inference, the temperature is set to 0 (i.e., greedy decoding) to reduce the variance. The CoT prompts contain $k=2, 4, 4, 8$ in-context examples for \textbf{TabMWP}, \textbf{GSM8K}, \textbf{Spider}, and \textbf{COGS}, respectively.

\subsection{Performance comparison results}
Table \ref{tab:results} presents experiment result summary. Detailed descriptions are  as follows:

\begin{table}[t]
\centering
\resizebox{1.0\columnwidth}{!}{%
\begin{tabular}{lcccc}
\toprule
 Method                    & TabMWP                  & GSM8K                     & Spider                    & COGS                          \\
\midrule
             \multicolumn{5}{c}{Backbone: gpt-3.5-turbo }\\
\midrule 
       Random                    & 62.4 \rise{0.0}~           & 75.7 \rise{0.0}              & 46.8 \rise{0.0}~           & 67.5 \rise{0.0}~                \\
                                    Retrieval-Q               & 72.3  \rise{9.9}~          & 75.6 \drop{0.1}           & 49.9 \rise{3.1}~           & 88.5 \rise{21.0}              \\
  %\texttt{Skill-KNN-large} & \textbf{78.3} \rise{15.9} & 75.0 \drop{0.7} & \textbf{58.4} \rise{11.6} & 94.6 \rise{27.2} \\  
  %\texttt{Skill-KNN-small} & 75.5 \rise{13.2} & 74.9 \drop{0.8} & 37.3 \drop{9.5}~ & 79.9 \rise{12.7} \\
  \texttt{Skill-KNN-zero} & 77.7 \rise{15.3} & 75.0 \drop{0.7} & 49.0 \rise{2.2}~ & 77.9 \rise{10.8} \\
  
                                     \cellcolor{gray!20} \rmethod  (ours)   &  \cellcolor{gray!20}\textbf{78.1} \rise{15.7} & \cellcolor{gray!20} \textbf{76.8} \rise{1.1} & \cellcolor{gray!20}\textbf{53.0} \rise{6.2}~ & \cellcolor{gray!20} \textbf{94.8} \rise{27.2}    \\ 
                                    \cmidrule{2-5}
                                     Retrieval-R (oracle)      & 77.4   \rise{15.0}       &  75.5 \drop{0.2}          & 64.4 \rise{17.6}         &   95.7 \rise{28.2}                        \\
\midrule
             \multicolumn{5}{c}{Backbone: gpt-4o }\\
\midrule 
Random                    & 87.6 \rise{0.0}~           & 78.1 \rise{0.0}            & 74.1 \rise{0.0}~               & 73.0 \rise{0.0}~               \\
                                     Retrieval-Q               & 85.9 \drop{1.7}~           &  78.1 \rise{0.0}          & 75.9 \rise{1.8}~  &   86.9 \rise{16.9}            \\
  %\texttt{Skill-KNN-large}& 80.6 \rise{11.3}& 62.0 \drop{0.2} & 56.3 \rise{9.8} & \textbf{96.8} \rise{23.4}\\
  %\texttt{Skill-KNN-small} & 77.4 \rise{8.1}~ & 62.3 \rise{0.1} & 47.4 \rise{0.3} & 79.4 \rise{6.0}~ \\
  \texttt{Skill-KNN-zero} & ~87.7 \rise{0.1}~ & \textbf{78.6} \drop{0.5} & 76.6 \rise{2.5} & 78.1 \rise{5.1}~ \\
                                    \cellcolor{gray!20}\rmethod (ours)     &\cellcolor{gray!20} \textbf{87.9} \rise{0.3} &\cellcolor{gray!20} 78.3 \rise{0.2}~ & \cellcolor{gray!20} \textbf{77.2} \rise{3.1}~          & \cellcolor{gray!20}  \textbf{90.2} \rise{17.2}  \\ 
                                    \cmidrule{2-5}
                                      Retrieval-R (oracle)     & 88.8   \rise{1.2}        & 77.1 \drop{1.0}          & 78.1 \rise{4.0}         &   92.8 \rise{19.8}           \\

\midrule
             \multicolumn{5}{c}{Backbone: claude-3-sonnet }\\
\midrule 
 Random                    &     92.6 \rise{0.0}      & 93.3 \rise{0.0}~            & 61.7 \rise{0.0}~             & 79.2 \rise{0.0}~            \\
                                     Retrieval-Q               &    93.1 \rise{0.5}   & 92.4 \drop{0.9}~           & 61.8 \rise{0.1}           & ~94.6 \rise{15.4}             \\
  %\texttt{Skill-KNN-large} & & 93.2 \drop{0.1} & \textbf{25.9} \rise{7.6} & 96.2 \rise{17.0} \\
  %\texttt{Skill-KNN-small} &  &  92.3 \drop{1.0} & 18.2 \drop{0.1} & 86.6 \rise{7.4} \\
  \texttt{Skill-KNN-zero} & 93.1 \rise{0.5} & 92.1 \drop{1.2}~ & 61.9 \rise{0.2} & 86.6 \rise{7.4}  \\
                                \cellcolor{gray!20} \rmethod (ours)     & \cellcolor{gray!20} \textbf{93.7} \rise{1.1}~  &\cellcolor{gray!20} \textbf{93.6} \rise{0.3}~ & \cellcolor{gray!20} \textbf{62.2} \rise{0.5}~ &\cellcolor{gray!20} ~\textbf{96.9} \rise{17.7}     \\ 
                                    \cmidrule{2-5}
                                     Retrieval-R (oracle)      &     94.1 \rise{1.5}    & 92.8 \drop{0.5}          & ~62.4 \rise{0.7}         & ~97.6 \rise{18.4}             \\
\midrule
             \multicolumn{5}{c}{Backbone: claude-3-haiku}\\
\midrule 
Random                    & 88.6 \rise{0.0} & 88.6 \rise{0.0} & 60.2 \rise{0.0} & 66.2 \rise{0.0}  \\
Retrieval-Q               &    92.2 \rise{3.6}   & ~88.6 \rise{0.0}~ & 60.0 \drop{0.2}           & ~88.5 \rise{22.3} \\
\texttt{Skill-KNN-zero} & 93.3 \rise{4.7} & ~\textbf{88.8} \rise{0.2}~ & 61.0 \rise{0.8} & ~79.7 \rise{13.5}  \\
\cellcolor{gray!20} \rmethod (ours)     & \cellcolor{gray!20} \textbf{93.3} \rise{4.7}~  &\cellcolor{gray!20} 87.6 \drop{1.0}~ & \cellcolor{gray!20} \textbf{61.3} \rise{1.1}~ &\cellcolor{gray!20} ~\textbf{89.9} \rise{23.7}     \\ 
\cmidrule{2-5}
Retrieval-R (oracle)      &     92.4 \rise{3.8}    & 88.9 \rise{0.3}          & ~61.2 \rise{1.0}         & ~96.5 \rise{30.3}             \\
\midrule
             \multicolumn{5}{c}{Backbone: Falcon-40B-Instruct }\\
\midrule 
Random                    & 45.7 \rise{0.0}~             & 38.8 \rise{0.0}             & 20.6 \rise{0.0}~            & 45.1 \rise{0.0}~               \\
                                    Retrieval-Q               & 51.9 \rise{6.2}~           & 37.3 \drop{1.5}           & 22.1 \rise{1.5}~           & 73.9 \rise{28.8}              \\
                                    %\texttt{Skill-KNN-large}                 & 55.9 \rise{10.2}           & \textbf{40.3} \rise{1.5} & 23.7 \rise{2.9} & 81.0 \rise{35.9}\\
                                    \texttt{Skill-KNN-small}                 & 51.4 \rise{5.7}~ & 36.5 \drop{2.3} & 20.3 \drop{0.3}~ & 59.4 \rise{14.3}   \\
                                    \texttt{Skill-KNN-zero} & 55.2 \rise{9.5}~ & 38.7 \drop{0.1} & 23.3 \rise{2.7} & 82.1 \rise{37.0}  \\
                                    \cellcolor{gray!20} \rmethod (ours)     &\cellcolor{gray!20} \textbf{57.7} \rise{12.0}& \cellcolor{gray!20}\textbf{39.1} \rise{0.3} & \cellcolor{gray!20}\textbf{24.8} \rise{4.2}~ &\cellcolor{gray!20} \textbf{89.5} \rise{44.4}    \\ 
                                    \cmidrule{2-5}
                                     Retrieval-R (oracle)      & 61.2 \rise{15.5}         & 40.4 \rise{1.6}          & 39.9 \rise{19.3}         & 90.3 \rise{45.2}             \\
\bottomrule
\end{tabular}}
\caption{Main results (\%) across all backbone models and datasets. Numbers in \textbf{bold} represent the best results for each backbone model across all selection methods. The subscripted gray values indicate the relative improvement over Random selection.}
\label{tab:results}
\vspace{-8pt}
\end{table}

\paragraph{\rmethod matches SOTA skill-based selection methods with superior computational efficiency. } \quad As shown in Table \ref{tab:results}, across all four benchmarks and five backbone models tested, \rmethod outperforms \texttt{Skill-KNN-zero} in 18 out of 20 experiments. This result highlights the effectiveness of the latent reasoning skills learned through unsupervised learning with small CVAE models, achieving comparable performance to the skill descriptions crafted by extensively pre-trained LLMs. Notably, \texttt{Skill-KNN-zero} uses the powerful LLM gpt-3.5-turbo for skill generations. However, in scenarios where only less capable LLMs are available, such as lacking an internet connection and requiring local inference, \texttt{Skill-KNN-small}, which uses the less capable LLM Falcon-40B-instruct, suffers significant performance drops across all four benchmarks. In contrast, \rmethod does not require powerful LLMs and achieves similar performance boosts for smaller backbone models like Falcon-40B-Instruct compared to \texttt{Skill-KNN-zero}.

Furthermore, in Table \ref{tab:compare_sample_efficiency}, we present a comparison of computational overhead, including computing time, estimated cost for pre-processing the example bank, and cost for each input query during selection, among \qmethod, \texttt{LaRS}, \texttt{Skill-KNN-zero}, and a supervised selection method PromptPG~\cite{Lu2022DynamicPL}. Our method achieves accuracy comparable to \texttt{Skill-KNN-zero}, requiring no LLM inferences (approximately \$30 savings per 1k examples) and reducing computing time by 1.5 hours per 1k examples during pre-processing, along with more than 100\% less cost per input query. Detailed experimental settings for estimating these costs can be found in Appendix \ref{appendix:expriments}. 
\paragraph{\rmethod is more robust to sub-optimal example banks.}~\texttt{Skill-KNN} selects examples based solely on the questions. For example, it selects examples whose questions require the same skills as the given question. However, sub-optimal example banks may include examples with incorrect or sub-optimal rationales, which should be avoided. In contrast, \rmethod considers both questions and rationales when computing the reasoning skill embedding, enhancing its robustness to sub-optimality. Table \ref{tab:suboptimal_bank} presents the answer accuracy of \texttt{Skill-KNN-zero} and \rmethod on the TabMWP and COGS benchmark with sub-optimal example banks, where 10\%, 20\% and 30\% of rationales are replaced by random rationales from the same example banks. \texttt{Skill-KNN-zero} suffers from a 3\% and 11.7\% performance drop at the replacement rate of 30\%, while \rmethod experiences only a 0.1\% and 1.9\% performance drop under the same conditions.
\vspace{-10pt}
\begin{table}
\centering
\resizebox{0.85\columnwidth}{!}{%
\begin{tabular}{lcccc}
\toprule
Benchmark             & \multicolumn{4}{c}{TabMWP}                                                                                             \\
Replace Rate (\%) & 0 & 0.1 & 0.2 & 0.3  \\
\midrule
%\texttt{Skill-KNN-large} &78.3 & 77.2 \drop{1.4\%} & 76.7 \drop{2.0\%} & 75.6 \drop{3.5\%}       \\
\texttt{Skill-KNN-zero}  &77.7 & 77.0 \drop{0.9\%} & 76.2 \drop{1.9\%} & 75.4 \drop{3.0\%}       \\
\rmethod                 &\textbf{78.1} & \textbf{78.1} \drop{0.0\%} & \textbf{78.0} \drop{0.1\%} & \textbf{77.9} \drop{0.1\%}       \\
\midrule
Benchmark             & \multicolumn{4}{c}{COGS}                                                                                             \\
Replace Rate (\%) & 0 & 0.1 & 0.2 & 0.3  \\
\midrule
%\texttt{Skill-KNN-large} &94.6 & 93.0 \drop{1.7\%} & 92.1 \drop{2.6\%} & 88.3 \drop{6.7\%}       \\
\texttt{Skill-KNN-zero}  &77.9 & 75.8 \drop{2.7\%} & 73.8 \drop{5.3\%} & 68.8 \drop{11.7\%}        \\
\rmethod                 &\textbf{94.8} & \textbf{94.7} \drop{0.1\%} & \textbf{93.3} \drop{1.6\%} & \textbf{93.0} \drop{1.9\%}    \\
\bottomrule
\end{tabular}%
}
\caption{Answer accuracy (\%) of \texttt{Skill-KNN-zero} and \rmethod~ on TabMWP and COGS benchmark with 0\%, 10\%, 20\%, and 30\% of the rationales in the example bank being replaced with random rationales. The subscripted gray values indicate the percentage drop relative to optimal example banks.\vspace{-8pt}}
\label{tab:suboptimal_bank}
\end{table}
\vspace{-6pt}

\begin{table}
\centering
\resizebox{\columnwidth}{!}
{%
\begin{tabular}{lclcccc}
        \toprule
          & \multirow{2}{*}{Accuracy (\%)$\uparrow$} & \multicolumn{2}{c}{Pre-processing} & Selection           \\
          &                                & Time (h/1k)$\downarrow$        & Cost (\$/1k)$\downarrow$        & Cost per query (\$)$\downarrow$ \\
         \midrule
         \method (ours) & \textbf{78.1} &  \textbf{0.5} \rise{0\%}& \textbf{\$0}~~ & \textbf{\$0.02} \rise{\%0}\\
         \texttt{Skill-KNN-zero} & 77.7 &  2 \rise{300\%} & \$30 & ~~~\$0.05 \rise{150\%} \\
         PromptPG & 74.2 &  6 \rise{1100\%}& \$50 & \$0.02 \rise{0\%}\\
         %EPR~\cite{Rubin2021LearningTR} &&  8& \\
         \midrule
         Retrieval-Q & 72.3 &  0 \drop{100\%} & \$0~~ & \$0.02 \rise{0\%}\\ 
         \bottomrule  
\end{tabular}%
}
    \caption{Comparison of accuracy and computational overhead, including computing time, estimated cost for pre-processing an example bank of 1k, and average cost per input query during selection, among four selection methods on the \textbf{TabMWP} dataset. The grey percentages represent the increased cost ratio associated with each selection method.}
    \label{tab:compare_sample_efficiency}
\end{table}
\section{Conclusions}
\vspace{-6pt}
This paper introduces \texttt{LaRS}, a novel demonstration selection method designed for CoT prompting. \method bases the selection on reasoning skills, which are latent representations discovered by unsupervised learning from rationales via a CVAE. Based on the experiments conducted across four LLMs and over four different reasoning tasks, \method manifests comparable performance on selecting effective few-shot examples for CoT reasoning while requiring no extra LLM inference and saving hours in pre-processing the example bank.

\section{Limitations}

Despite the success of \texttt{LaRS}, a few limitations and potential future directions are worth noting. First, the impact of the order of examples in the prompts is not considered. Introducing additional heuristics to sort the examples could potentially lead to better performances. Second, in the CVAE, the decoder is represented by an MLP neural network. However, it would be ideal to represent the decoder as a prompt-tuning module, which aligns better with the implicit skill model assumption. Finally, one single reasoning skill might not be sufficient to represent the entire rationale that might contain multiple steps of reasoning. Learning and selecting reasoning skills for each individual reasoning step is an interesting direction to explore.

\newpage
\newpage
\bibliography{custom}

\begin{thebibliography}{39}
\expandafter\ifx\csname natexlab\endcsname\relax\def\natexlab#1{#1}\fi

\bibitem[{An et~al.(2023{\natexlab{a}})An, Lin, Fu, Chen, Zheng, Lou, and Zhang}]{An2023HowDI}
Shengnan An, Zeqi Lin, Qiang Fu, B.~Chen, Nanning Zheng, Jian-Guang Lou, and D.~Zhang. 2023{\natexlab{a}}.
\newblock \href {https://api.semanticscholar.org/CorpusID:258558112} {How do in-context examples affect compositional generalization?}
\newblock \emph{ArXiv}, abs/2305.04835.

\bibitem[{An et~al.(2023{\natexlab{b}})An, Zhou, Lin, Fu, Chen, Zheng, Chen, and Lou}]{an2023skill}
Shengnan An, Bo~Zhou, Zeqi Lin, Qiang Fu, Bei Chen, Nanning Zheng, Weizhu Chen, and Jian-Guang Lou. 2023{\natexlab{b}}.
\newblock Skill-based few-shot selection for in-context learning.
\newblock \emph{arXiv preprint arXiv:2305.14210}.

\bibitem[{Bommasani et~al.(2021)Bommasani, Hudson, Adeli, Altman, Arora, von Arx, Bernstein, Bohg, Bosselut, Brunskill, Brynjolfsson, Buch, Card, Castellon, Chatterji, Chen, Creel, Davis, Demszky, Donahue, Doumbouya, Durmus, Ermon, Etchemendy, Ethayarajh, Fei-Fei, Finn, Gale, Gillespie, Goel, Goodman, Grossman, Guha, Hashimoto, Henderson, Hewitt, Ho, Hong, Hsu, Huang, Icard, Jain, Jurafsky, Kalluri, Karamcheti, Keeling, Khani, Khattab, Koh, Krass, Krishna, Kuditipudi, Kumar, Ladhak, Lee, Lee, Leskovec, Levent, Li, Li, Ma, Malik, Manning, Mirchandani, Mitchell, Munyikwa, Nair, Narayan, Narayanan, Newman, Nie, Niebles, Nilforoshan, Nyarko, Ogut, Orr, Papadimitriou, Park, Piech, Portelance, Potts, Raghunathan, Reich, Ren, Rong, Roohani, Ruiz, Ryan, R'e, Sadigh, Sagawa, Santhanam, Shih, Srinivasan, Tamkin, Taori, Thomas, Tram{\`e}r, Wang, Wang, Wu, Wu, Wu, Xie, Yasunaga, You, Zaharia, Zhang, Zhang, Zhang, Zhang, Zheng, Zhou, and Liang}]{Bommasani2021OnTO}
Rishi Bommasani, Drew~A. Hudson, Ehsan Adeli, Russ Altman, Simran Arora, Sydney von Arx, Michael~S. Bernstein, Jeannette Bohg, Antoine Bosselut, Emma Brunskill, Erik Brynjolfsson, S.~Buch, Dallas Card, Rodrigo Castellon, Niladri~S. Chatterji, Annie~S. Chen, Kathleen~A. Creel, Jared Davis, Dora Demszky, Chris Donahue, Moussa Doumbouya, Esin Durmus, Stefano Ermon, John Etchemendy, Kawin Ethayarajh, Li~Fei-Fei, Chelsea Finn, Trevor Gale, Lauren~E. Gillespie, Karan Goel, Noah~D. Goodman, Shelby Grossman, Neel Guha, Tatsunori Hashimoto, Peter Henderson, John Hewitt, Daniel~E. Ho, Jenny Hong, Kyle Hsu, Jing Huang, Thomas~F. Icard, Saahil Jain, Dan Jurafsky, Pratyusha Kalluri, Siddharth Karamcheti, Geoff Keeling, Fereshte Khani, O.~Khattab, Pang~Wei Koh, Mark~S. Krass, Ranjay Krishna, Rohith Kuditipudi, Ananya Kumar, Faisal Ladhak, Mina Lee, Tony Lee, Jure Leskovec, Isabelle Levent, Xiang~Lisa Li, Xuechen Li, Tengyu Ma, Ali Malik, Christopher~D. Manning, Suvir Mirchandani, Eric Mitchell, Zanele Munyikwa, Suraj Nair,
  Avanika Narayan, Deepak Narayanan, Benjamin Newman, Allen Nie, Juan~Carlos Niebles, Hamed Nilforoshan, J.~F. Nyarko, Giray Ogut, Laurel~J. Orr, Isabel Papadimitriou, Joon~Sung Park, Chris Piech, Eva Portelance, Christopher Potts, Aditi Raghunathan, Robert Reich, Hongyu Ren, Frieda Rong, Yusuf~H. Roohani, Camilo Ruiz, Jack Ryan, Christopher R'e, Dorsa Sadigh, Shiori Sagawa, Keshav Santhanam, Andy Shih, Krishna~Parasuram Srinivasan, Alex Tamkin, Rohan Taori, Armin~W. Thomas, Florian Tram{\`e}r, Rose~E. Wang, William Wang, Bohan Wu, Jiajun Wu, Yuhuai Wu, Sang~Michael Xie, Michihiro Yasunaga, Jiaxuan You, Matei~A. Zaharia, Michael Zhang, Tianyi Zhang, Xikun Zhang, Yuhui Zhang, Lucia Zheng, Kaitlyn Zhou, and Percy Liang. 2021.
\newblock On the opportunities and risks of foundation models.
\newblock \emph{ArXiv}, abs/2108.07258.

\bibitem[{Brown et~al.(2020)Brown, Mann, Ryder, Subbiah, Kaplan, Dhariwal, Neelakantan, Shyam, Sastry, Askell, Agarwal, Herbert-Voss, Krueger, Henighan, Child, Ramesh, Ziegler, Wu, Winter, Hesse, Chen, Sigler, Litwin, Gray, Chess, Clark, Berner, McCandlish, Radford, Sutskever, and Amodei}]{Brown2020LanguageMA}
Tom~B. Brown, Benjamin Mann, Nick Ryder, Melanie Subbiah, Jared Kaplan, Prafulla Dhariwal, Arvind Neelakantan, Pranav Shyam, Girish Sastry, Amanda Askell, Sandhini Agarwal, Ariel Herbert-Voss, Gretchen Krueger, T.~J. Henighan, Rewon Child, Aditya Ramesh, Daniel~M. Ziegler, Jeff Wu, Clemens Winter, Christopher Hesse, Mark Chen, Eric Sigler, Mateusz Litwin, Scott Gray, Benjamin Chess, Jack Clark, Christopher Berner, Sam McCandlish, Alec Radford, Ilya Sutskever, and Dario Amodei. 2020.
\newblock Language models are few-shot learners.
\newblock \emph{ArXiv}, abs/2005.14165.

\bibitem[{Chen et~al.(2023)Chen, Pan, Yu, Song, Wang, Yu, and Chen}]{chen2023skills}
Jiaao Chen, Xiaoman Pan, Dian Yu, Kaiqiang Song, Xiaoyang Wang, Dong Yu, and Jianshu Chen. 2023.
\newblock Skills-in-context prompting: Unlocking compositionality in large language models.
\newblock \emph{arXiv preprint arXiv:2308.00304}.

\bibitem[{Chowdhery et~al.(2022)Chowdhery, Narang, Devlin, Bosma, Mishra, Roberts, Barham, Chung, Sutton, Gehrmann, Schuh, Shi, Tsvyashchenko, Maynez, Rao, Barnes, Tay, Shazeer, Prabhakaran, Reif, Du, Hutchinson, Pope, Bradbury, Austin, Isard, Gur-Ari, Yin, Duke, Levskaya, Ghemawat, Dev, Michalewski, Garc{\'i}a, Misra, Robinson, Fedus, Zhou, Ippolito, Luan, Lim, Zoph, Spiridonov, Sepassi, Dohan, Agrawal, Omernick, Dai, Pillai, Pellat, Lewkowycz, Moreira, Child, Polozov, Lee, Zhou, Wang, Saeta, D{\'i}az, Firat, Catasta, Wei, Meier-Hellstern, Eck, Dean, Petrov, and Fiedel}]{Chowdhery2022PaLMSL}
Aakanksha Chowdhery, Sharan Narang, Jacob Devlin, Maarten Bosma, Gaurav Mishra, Adam Roberts, Paul Barham, Hyung~Won Chung, Charles Sutton, Sebastian Gehrmann, Parker Schuh, Kensen Shi, Sasha Tsvyashchenko, Joshua Maynez, Abhishek Rao, Parker Barnes, Yi~Tay, Noam~M. Shazeer, Vinodkumar Prabhakaran, Emily Reif, Nan Du, Benton~C. Hutchinson, Reiner Pope, James Bradbury, Jacob Austin, Michael Isard, Guy Gur-Ari, Pengcheng Yin, Toju Duke, Anselm Levskaya, Sanjay Ghemawat, Sunipa Dev, Henryk Michalewski, Xavier Garc{\'i}a, Vedant Misra, Kevin Robinson, Liam Fedus, Denny Zhou, Daphne Ippolito, David Luan, Hyeontaek Lim, Barret Zoph, Alexander Spiridonov, Ryan Sepassi, David Dohan, Shivani Agrawal, Mark Omernick, Andrew~M. Dai, Thanumalayan~Sankaranarayana Pillai, Marie Pellat, Aitor Lewkowycz, Erica Moreira, Rewon Child, Oleksandr Polozov, Katherine Lee, Zongwei Zhou, Xuezhi Wang, Brennan Saeta, Mark D{\'i}az, Orhan Firat, Michele Catasta, Jason Wei, Kathleen~S. Meier-Hellstern, Douglas Eck, Jeff Dean, Slav Petrov,
  and Noah Fiedel. 2022.
\newblock Palm: Scaling language modeling with pathways.
\newblock \emph{ArXiv}, abs/2204.02311.

\bibitem[{Cobbe et~al.(2021{\natexlab{a}})Cobbe, Kosaraju, Bavarian, Chen, Jun, Kaiser, Plappert, Tworek, Hilton, Nakano, Hesse, and Schulman}]{Cobbe2021TrainingVT}
Karl Cobbe, Vineet Kosaraju, Mohammad Bavarian, Mark Chen, Heewoo Jun, Lukasz Kaiser, Matthias Plappert, Jerry Tworek, Jacob Hilton, Reiichiro Nakano, Christopher Hesse, and John Schulman. 2021{\natexlab{a}}.
\newblock Training verifiers to solve math word problems.
\newblock \emph{ArXiv}, abs/2110.14168.

\bibitem[{Cobbe et~al.(2021{\natexlab{b}})Cobbe, Kosaraju, Bavarian, Chen, Jun, Kaiser, Plappert, Tworek, Hilton, Nakano et~al.}]{cobbe2021training}
Karl Cobbe, Vineet Kosaraju, Mohammad Bavarian, Mark Chen, Heewoo Jun, Lukasz Kaiser, Matthias Plappert, Jerry Tworek, Jacob Hilton, Reiichiro Nakano, et~al. 2021{\natexlab{b}}.
\newblock Training verifiers to solve math word problems.
\newblock \emph{arXiv preprint arXiv:2110.14168}.

\bibitem[{Devlin et~al.(2019)Devlin, Chang, Lee, and Toutanova}]{Devlin2019BERTPO}
Jacob Devlin, Ming-Wei Chang, Kenton Lee, and Kristina Toutanova. 2019.
\newblock Bert: Pre-training of deep bidirectional transformers for language understanding.
\newblock \emph{ArXiv}, abs/1810.04805.

\bibitem[{Diao et~al.(2023)Diao, Wang, Lin, and Zhang}]{Diao2023ActivePW}
Shizhe Diao, Pengcheng Wang, Yong Lin, and Tong Zhang. 2023.
\newblock Active prompting with chain-of-thought for large language models.
\newblock \emph{ArXiv}, abs/2302.12246.

\bibitem[{Dong et~al.(2022)Dong, Li, Dai, Zheng, Wu, Chang, Sun, Xu, and Sui}]{dong2022survey}
Qingxiu Dong, Lei Li, Damai Dai, Ce~Zheng, Zhiyong Wu, Baobao Chang, Xu~Sun, Jingjing Xu, and Zhifang Sui. 2022.
\newblock A survey for in-context learning.
\newblock \emph{arXiv preprint arXiv:2301.00234}.

\bibitem[{Gao et~al.(2021)Gao, Fisch, and Chen}]{Gao2021MakingPL}
Tianyu Gao, Adam Fisch, and Danqi Chen. 2021.
\newblock Making pre-trained language models better few-shot learners.
\newblock \emph{ArXiv}, abs/2012.15723.

\bibitem[{Gupta et~al.(2023)Gupta, Singh, and Gardner}]{Gupta2023CoveragebasedES}
Shivanshu Gupta, Sameer Singh, and Matt Gardner. 2023.
\newblock Coverage-based example selection for in-context learning.
\newblock \emph{ArXiv}, abs/2305.14907.

\bibitem[{He et~al.(2021)He, Liu, Gao, and Chen}]{he2021deberta}
Pengcheng He, Xiaodong Liu, Jianfeng Gao, and Weizhu Chen. 2021.
\newblock \href {https://openreview.net/forum?id=XPZIaotutsD} {Deberta: Decoding-enhanced bert with disentangled attention}.
\newblock In \emph{International Conference on Learning Representations}.

\bibitem[{Hu et~al.(2022)Hu, Lee, Xie, Yu, Smith, and Ostendorf}]{Hu2022InContextLF}
Yushi Hu, Chia-Hsuan Lee, Tianbao Xie, Tao Yu, Noah~A. Smith, and Mari Ostendorf. 2022.
\newblock In-context learning for few-shot dialogue state tracking.
\newblock \emph{ArXiv}, abs/2203.08568.

\bibitem[{Kim and Linzen(2020)}]{Kim2020COGSAC}
Najoung Kim and Tal Linzen. 2020.
\newblock \href {https://api.semanticscholar.org/CorpusID:222290851} {Cogs: A compositional generalization challenge based on semantic interpretation}.
\newblock \emph{ArXiv}, abs/2010.05465.

\bibitem[{Kojima et~al.(2022)Kojima, Gu, Reid, Matsuo, and Iwasawa}]{kojima2022large}
Takeshi Kojima, Shixiang~Shane Gu, Machel Reid, Yutaka Matsuo, and Yusuke Iwasawa. 2022.
\newblock Large language models are zero-shot reasoners.
\newblock \emph{Advances in neural information processing systems}, 35:22199--22213.

\bibitem[{Liu et~al.(2021)Liu, Shen, Zhang, Dolan, Carin, and Chen}]{Liu2021WhatMG}
Jiachang Liu, Dinghan Shen, Yizhe Zhang, Bill Dolan, Lawrence Carin, and Weizhu Chen. 2021.
\newblock What makes good in-context examples for gpt-3?
\newblock In \emph{Workshop on Knowledge Extraction and Integration for Deep Learning Architectures; Deep Learning Inside Out}.

\bibitem[{Lu et~al.(2022)Lu, Qiu, Chang, Wu, Zhu, Rajpurohit, Clark, and Kalyan}]{Lu2022DynamicPL}
Pan Lu, Liang Qiu, Kai-Wei Chang, Ying~Nian Wu, Song-Chun Zhu, Tanmay Rajpurohit, Peter Clark, and A.~Kalyan. 2022.
\newblock \href {https://api.semanticscholar.org/CorpusID:252595921} {Dynamic prompt learning via policy gradient for semi-structured mathematical reasoning}.
\newblock \emph{ArXiv}, abs/2209.14610.

\bibitem[{Lu et~al.(2021)Lu, Bartolo, Moore, Riedel, and Stenetorp}]{Lu2021FantasticallyOP}
Yao Lu, Max Bartolo, Alastair Moore, Sebastian Riedel, and Pontus Stenetorp. 2021.
\newblock Fantastically ordered prompts and where to find them: Overcoming few-shot prompt order sensitivity.
\newblock \emph{ArXiv}, abs/2104.08786.

\bibitem[{Min et~al.(2022)Min, Lyu, Holtzman, Artetxe, Lewis, Hajishirzi, and Zettlemoyer}]{min2022rethinking}
Sewon Min, Xinxi Lyu, Ari Holtzman, Mikel Artetxe, Mike Lewis, Hannaneh Hajishirzi, and Luke Zettlemoyer. 2022.
\newblock Rethinking the role of demonstrations: What makes in-context learning work?
\newblock \emph{arXiv preprint arXiv:2202.12837}.

\bibitem[{Neelakantan et~al.(2022)Neelakantan, Xu, Puri, Radford, Han, Tworek, Yuan, Tezak, Kim, Hallacy et~al.}]{neelakantan2022text}
Arvind Neelakantan, Tao Xu, Raul Puri, Alec Radford, Jesse~Michael Han, Jerry Tworek, Qiming Yuan, Nikolas Tezak, Jong~Wook Kim, Chris Hallacy, et~al. 2022.
\newblock Text and code embeddings by contrastive pre-training.
\newblock \emph{arXiv preprint arXiv:2201.10005}.

\bibitem[{Nye et~al.(2021)Nye, Andreassen, Gur-Ari, Michalewski, Austin, Bieber, Dohan, Lewkowycz, Bosma, Luan, Sutton, and Odena}]{Nye2021ShowYW}
Maxwell Nye, Anders Andreassen, Guy Gur-Ari, Henryk Michalewski, Jacob Austin, David Bieber, David Dohan, Aitor Lewkowycz, Maarten Bosma, David Luan, Charles Sutton, and Augustus Odena. 2021.
\newblock Show your work: Scratchpads for intermediate computation with language models.
\newblock \emph{ArXiv}, abs/2112.00114.

\bibitem[{Rae et~al.(2021)Rae, Borgeaud, Cai, Millican, Hoffmann, Song, Aslanides, Henderson, Ring, Young, Rutherford, Hennigan, Menick, Cassirer, Powell, van~den Driessche, Hendricks, Rauh, Huang, Glaese, Welbl, Dathathri, Huang, Uesato, Mellor, Higgins, Creswell, McAleese, Wu, Elsen, Jayakumar, Buchatskaya, Budden, Sutherland, Simonyan, Paganini, Sifre, Martens, Li, Kuncoro, Nematzadeh, Gribovskaya, Donato, Lazaridou, Mensch, Lespiau, Tsimpoukelli, Grigorev, Fritz, Sottiaux, Pajarskas, Pohlen, Gong, Toyama, de~Masson~d'Autume, Li, Terzi, Mikulik, Babuschkin, Clark, de~Las~Casas, Guy, Jones, Bradbury, Johnson, Hechtman, Weidinger, Gabriel, Isaac, Lockhart, Osindero, Rimell, Dyer, Vinyals, Ayoub, Stanway, Bennett, Hassabis, Kavukcuoglu, and Irving}]{Rae2021ScalingLM}
Jack~W. Rae, Sebastian Borgeaud, Trevor Cai, Katie Millican, Jordan Hoffmann, Francis Song, John Aslanides, Sarah Henderson, Roman Ring, Susannah Young, Eliza Rutherford, Tom Hennigan, Jacob Menick, Albin Cassirer, Richard Powell, George van~den Driessche, Lisa~Anne Hendricks, Maribeth Rauh, Po-Sen Huang, Amelia Glaese, Johannes Welbl, Sumanth Dathathri, Saffron Huang, Jonathan Uesato, John F.~J. Mellor, Irina Higgins, Antonia Creswell, Nathan McAleese, Amy Wu, Erich Elsen, Siddhant~M. Jayakumar, Elena Buchatskaya, David Budden, Esme Sutherland, Karen Simonyan, Michela Paganini, L.~Sifre, Lena Martens, Xiang~Lorraine Li, Adhiguna Kuncoro, Aida Nematzadeh, Elena Gribovskaya, Domenic Donato, Angeliki Lazaridou, Arthur Mensch, Jean-Baptiste Lespiau, Maria Tsimpoukelli, N.~K. Grigorev, Doug Fritz, Thibault Sottiaux, Mantas Pajarskas, Tobias Pohlen, Zhitao Gong, Daniel Toyama, Cyprien de~Masson~d'Autume, Yujia Li, Tayfun Terzi, Vladimir Mikulik, Igor Babuschkin, Aidan Clark, Diego de~Las~Casas, Aurelia Guy, Chris
  Jones, James Bradbury, Matthew~G. Johnson, Blake~A. Hechtman, Laura Weidinger, Iason Gabriel, William~S. Isaac, Edward Lockhart, Simon Osindero, Laura Rimell, Chris Dyer, Oriol Vinyals, Kareem~W. Ayoub, Jeff Stanway, L.~L. Bennett, Demis Hassabis, Koray Kavukcuoglu, and Geoffrey Irving. 2021.
\newblock Scaling language models: Methods, analysis \& insights from training gopher.
\newblock \emph{ArXiv}, abs/2112.11446.

\bibitem[{Reimers and Gurevych(2019)}]{reimers2019sentence}
Nils Reimers and Iryna Gurevych. 2019.
\newblock Sentence-bert: Sentence embeddings using siamese bert-networks.
\newblock \emph{arXiv preprint arXiv:1908.10084}.

\bibitem[{Saunshi et~al.(2020)Saunshi, Malladi, and Arora}]{Saunshi2020AME}
Nikunj Saunshi, Sadhika Malladi, and Sanjeev Arora. 2020.
\newblock \href {https://api.semanticscholar.org/CorpusID:222208920} {A mathematical exploration of why language models help solve downstream tasks}.
\newblock \emph{ArXiv}, abs/2010.03648.

\bibitem[{Shin et~al.(2020)Shin, Razeghi, IV, Wallace, and Singh}]{Shin2020ElicitingKF}
Taylor Shin, Yasaman Razeghi, Robert L~Logan IV, Eric Wallace, and Sameer Singh. 2020.
\newblock \href {https://api.semanticscholar.org/CorpusID:226222232} {Eliciting knowledge from language models using automatically generated prompts}.
\newblock In \emph{Conference on Empirical Methods in Natural Language Processing}.

\bibitem[{Suzgun et~al.(2022)Suzgun, Scales, Scharli, Gehrmann, Tay, Chung, Chowdhery, Le, hsin Chi, Zhou, and Wei}]{Suzgun2022ChallengingBT}
Mirac Suzgun, Nathan Scales, Nathanael Scharli, Sebastian Gehrmann, Yi~Tay, Hyung~Won Chung, Aakanksha Chowdhery, Quoc~V. Le, Ed~Huai hsin Chi, Denny Zhou, and Jason Wei. 2022.
\newblock Challenging big-bench tasks and whether chain-of-thought can solve them.
\newblock In \emph{Annual Meeting of the Association for Computational Linguistics}.

\bibitem[{Vaswani et~al.(2017)Vaswani, Shazeer, Parmar, Uszkoreit, Jones, Gomez, Kaiser, and Polosukhin}]{Vaswani2017AttentionIA}
Ashish Vaswani, Noam~M. Shazeer, Niki Parmar, Jakob Uszkoreit, Llion Jones, Aidan~N. Gomez, Lukasz Kaiser, and Illia Polosukhin. 2017.
\newblock Attention is all you need.
\newblock In \emph{NIPS}.

\bibitem[{Wang et~al.(2022)Wang, Min, Deng, Shen, Wu, Zettlemoyer, and Sun}]{wang2022towards}
Boshi Wang, Sewon Min, Xiang Deng, Jiaming Shen, You Wu, Luke Zettlemoyer, and Huan Sun. 2022.
\newblock Towards understanding chain-of-thought prompting: An empirical study of what matters.
\newblock \emph{arXiv preprint arXiv:2212.10001}.

\bibitem[{Wang et~al.(2023)Wang, Zhu, and Wang}]{wang2023large}
Xinyi Wang, Wanrong Zhu, and William~Yang Wang. 2023.
\newblock Large language models are implicitly topic models: Explaining and finding good demonstrations for in-context learning.
\newblock \emph{arXiv preprint arXiv:2301.11916}.

\bibitem[{Wei et~al.(2021)Wei, Xie, and Ma}]{Wei2021WhyDP}
Colin Wei, Sang~Michael Xie, and Tengyu Ma. 2021.
\newblock Why do pretrained language models help in downstream tasks? an analysis of head and prompt tuning.
\newblock \emph{ArXiv}, abs/2106.09226.

\bibitem[{Wei et~al.(2022{\natexlab{a}})Wei, Tay, Bommasani, Raffel, Zoph, Borgeaud, Yogatama, Bosma, Zhou, Metzler, hsin Chi, Hashimoto, Vinyals, Liang, Dean, and Fedus}]{Wei2022EmergentAO}
Jason Wei, Yi~Tay, Rishi Bommasani, Colin Raffel, Barret Zoph, Sebastian Borgeaud, Dani Yogatama, Maarten Bosma, Denny Zhou, Donald Metzler, Ed~Huai hsin Chi, Tatsunori Hashimoto, Oriol Vinyals, Percy Liang, Jeff Dean, and William Fedus. 2022{\natexlab{a}}.
\newblock Emergent abilities of large language models.
\newblock \emph{Trans. Mach. Learn. Res.}, 2022.

\bibitem[{Wei et~al.(2022{\natexlab{b}})Wei, Wang, Schuurmans, Bosma, hsin Chi, Xia, Le, and Zhou}]{Wei2022ChainOT}
Jason Wei, Xuezhi Wang, Dale Schuurmans, Maarten Bosma, Ed~Huai hsin Chi, F.~Xia, Quoc Le, and Denny Zhou. 2022{\natexlab{b}}.
\newblock Chain of thought prompting elicits reasoning in large language models.
\newblock \emph{ArXiv}, abs/2201.11903.

\bibitem[{Xie et~al.(2021{\natexlab{a}})Xie, Raghunathan, Liang, and Ma}]{Xie2021AnEO}
Sang~Michael Xie, Aditi Raghunathan, Percy Liang, and Tengyu Ma. 2021{\natexlab{a}}.
\newblock An explanation of in-context learning as implicit bayesian inference.
\newblock \emph{ArXiv}, abs/2111.02080.

\bibitem[{Xie et~al.(2021{\natexlab{b}})Xie, Raghunathan, Liang, and Ma}]{xie2021explanation}
Sang~Michael Xie, Aditi Raghunathan, Percy Liang, and Tengyu Ma. 2021{\natexlab{b}}.
\newblock An explanation of in-context learning as implicit bayesian inference.
\newblock \emph{arXiv preprint arXiv:2111.02080}.

\bibitem[{Xu et~al.(2023)Xu, Guo, Duan, and McAuley}]{xu2023baize}
Canwen Xu, Daya Guo, Nan Duan, and Julian McAuley. 2023.
\newblock Baize: An open-source chat model with parameter-efficient tuning on self-chat data.
\newblock \emph{arXiv preprint arXiv:2304.01196}.

\bibitem[{Yu et~al.(2018)Yu, Zhang, Yang, Yasunaga, Wang, Li, Ma, Li, Yao, Roman, Zhang, and Radev}]{Yu2018SpiderAL}
Tao Yu, Rui Zhang, Kai-Chou Yang, Michihiro Yasunaga, Dongxu Wang, Zifan Li, James Ma, Irene~Z Li, Qingning Yao, Shanelle Roman, Zilin Zhang, and Dragomir~R. Radev. 2018.
\newblock \href {https://api.semanticscholar.org/CorpusID:52815560} {Spider: A large-scale human-labeled dataset for complex and cross-domain semantic parsing and text-to-sql task}.
\newblock \emph{ArXiv}, abs/1809.08887.

\bibitem[{Zhang et~al.(2022)Zhang, Zhang, Li, and Smola}]{Zhang2022AutomaticCO}
Zhuosheng Zhang, Aston Zhang, Mu~Li, and Alexander~J. Smola. 2022.
\newblock Automatic chain of thought prompting in large language models.
\newblock \emph{ArXiv}, abs/2210.03493.

\end{thebibliography}

\onecolumn
\appendix
\newpage

\pagestyle{plain}
%\setcounter{page}{0}
%    \pagenumbering{arabic}
%    \setcounter{page}{1}
    
\begin{center}
\linespread{1.1} 
%\Huge {\bf Appendix}\\
\huge \bf Appendix: LaRS: Latent Reasoning Skill for Chain-of-Thought Reasoning
\end{center}

\section{LaRS Demonstration Selection}
A practical desmonstration selection process for LaRS that tackle the difficulty of sampling from an unknown distribution $P_E(Q, \RA \mid z)$ is described as follows. To begin with, LaRS learns reasoning skill encoder $\pi_\phi$ and reasoning policy $q_\omega$. For a given test question $\Qt$, the desirable reasoning skill $z_{\text{test}} = \argmax_z [\pi_{\phi}(z|f(Q_{\text{test}}))]$ can be computed using the reasoning policy. Subsequently, each example from the example bank can be scored based on the cosine similarity between $z_{\text{test}}$ and $z_{\text{post}}$, where $z_{\text{post}} = \argmax_z [q_{\omega}(z|Q,\RA))]$ represents the maximum likelihood skill of the current example. Finally, a CoT prompt can be constructed by selecting the top-$k$ examples according to the computed scores. The step-by-step procedure is outlined in Algorithm \ref{alg:retrieval_RSD}.
\begin{algorithm}
\caption{Demonstration selection}
\label{alg:retrieval_RSD}
\textbf{Input}: Test question $Q_{\text{test}}$, a pre-trained embedding model $f$, a reasoning policy $\pi_\phi(z|f(Q))$, a reasoning skill encoder $q_\omega(z|f(Q, \RA))$, and an example bank $\mathcal{D}_E = \{(Q^j, \RA^j)\}_j$. \\
\textbf{Parameter}: shot number $k$ \\
\textbf{Output}: $(Q_1, \RA_1, Q_2, \RA_2, \cdots, Q_k, \RA_k)$\\
\begin{algorithmic}[1] %[1] enables line numbers
\STATE Compute $z_{\text{test}} \leftarrow$ mean of $\pi(z|f(Q_{\text{test}}))$
\FOR{each $(Q^j, \RA^j)$ in $\mathcal{D}_E$}
\STATE Compute $z^j_{\text{post}} \leftarrow$ mean of $q_\omega(z|f(Q^j, \RA^j))$
\STATE Compute $r^j = \frac{z_{\text{test}} \cdot {z^j_{\text{post}}}^\intercal}{|z_{\text{test}} | \cdot |z^j_{\text{post}}|}$
\ENDFOR
\STATE Select top-$k$ demonstrations with the largest $r^j$ and sort them in ascending order, denoted as $(Q_1, \RA_1, Q_2, \RA_2, \cdots, Q_k, \RA_k)$.
\RETURN $(Q_1, \RA_1, Q_2, \RA_2, \cdots, Q_k, \RA_k)$
\end{algorithmic}
\end{algorithm}

\section{Experimental Details}
\label{appendix:expriments}
\subsection{Dataset}
We provide a detailed description of the dataset and the split of train and test set as follows:
\paragraph{TabMWP}~\cite{Lu2022DynamicPL} This dataset consists of semi-structured mathematical reasoning problems, comprising 38,431 open-domain grade-level problems that require mathematical reasoning on both textual and tabular data. We use the train set, containing 23,059 examples, to train our \method models, and test1k set containing 1K examples to evaluate the selection methods.
\paragraph{Spider}~\cite{Yu2018SpiderAL} Spider is a large-scale text-to-SQL dataset. It includes a train set with 7,000 examples and a dev set with 1,034 examples. We use the train set to train our \method models, and the dev set as the test set to evaluate the selection methods.
\paragraph{COGS}~\cite{Kim2020COGSAC} is a synthetic benchmark for testing compositional generalization in semantic parsing. We transform the output format in the same way as \citet{An2023HowDI}, and consider a mixture of two sub-tasks: primitive substitution (P.S.) and primitive structural alternation (P.A.). This results in a train set of 6916 examples to train our \method models and a test set of 1000 examples to evaluate the selection method.
\paragraph{GSM8k}~\cite{cobbe2021training} GSM8k is a dataset containing 8.5K high-quality, linguistically diverse grade school math word problems. It includes a train set of 7.5K problems and a test set of 1319 problems. We use the train set to train our \method models, and the test set to evaluate the selection methods.

\subsection{LaRS Implementation Details}
\method contains a encoder, a decoder, and a reasoning policy model. The reasoning skill is represented as a 128-dimensional continuous space. Both the encoder and the reasoning policy model are represented as a feed-forward multiple layer perception (MLP) with two 256-unit hidden layers, predicting the mean and variance of a multivariate Gaussian distribution in the latent space of reasoning skills. The decoder is a MLP with two 256-unit hidden layers that predicts a value in the embedding space deterministically. The dimension of the embedding space depends on the choice of pre-trained embedding models. The models are trained using the loss function in Equation~\ref{eq:loss} with a batch size of 256 and a learning rate of 0.0001 for 1000 epochs on a machine with 48 CPU cores and a Nvidia A40 GPU. Those hyper-parameters apply for all four datasets.

\subsection{Skill-KNN Implementation Details}
We used the same skill annotations as the original \texttt{Skill-KNN} implementation for COGS and Spider dataset. For TabMWP and GSM8K, we manually create skill annotations for 8 questions for each dataset. The new skill annotations are shown in Table \ref{tab:skill_label_tab} and \ref{tab:skill_label_gsm8k}.
\begin{table}[t]
\centering
\resizebox{\columnwidth}{!}{%
    \begin{tabular}{c p{4cm} p{6cm} p{10cm}}
    \toprule
    ID & Table & Question & Skill Description \\
    \midrule
    1 & Name | Score\newline Jackson | 32\newline Madelyn | 31\newline Gary | 36\newline Suzie | 33\newline Edgar | 31\newline Ben | 32\newline Felipe | 29 & Some friends played miniature golf and wrote down their scores. What is the range of the numbers? & To solve this problem, we need to find the greatest number and the least number. Then, subtract the least number from the greatest number.  \\
    \midrule
    2 & x | y\newline 17 | 13\newline 18 | 6\newline 19 | 2 & The table shows a function. Is the function linear or nonlinear? & To solve this problem, we need to compare the rate of change between any two rows of the table. \\     \midrule
    3 & box of tissues | \$0.90\newline of hand lotion | \$0.94\newline tube of toothpaste | \$0.84\newline package of dental floss | \$0.85\newline box of bandages | \$0.87\newline bottle of nail polish | \$0.99 & Sophie has \$1.50. Does she have enough to buy a box of tissues and a package of dental floss? & To solve this problem, we need to compute the total cost and compare it with the budget. \\
    \midrule
    4 & Day | Number of fan letters\newline Monday | 3,985\newline Tuesday | 1,207\newline Wednesday | 6,479\newline Thursday | 2,715\newline Friday | 8,078 & An actor was informed how many fan letters he received each day. How many more fan letters were received on Friday than on Tuesday? & To solve the problem, we need to locate the two values in the table and do subtraction. \\
    \midrule
    5 & Stem | Leaf \newline 3 | 1, 5, 7, 8\newline 4 | 0, 3, 5, 5, 8\newline 5 | 2, 4, 5, 7, 9\newline 6 | 4, 5, 6\newline 7 | 1, 1, 7, 8\newline 8 | \newline 9 | 0 & Daniel counted the number of silver beads on each bracelet at Lowell Jewelry, the store where he works. What is the largest number of silver beads? & To solve this problem, we need to locate the largest number from a stem-and-leaf plot. \\
    \midrule
    6 & Number of tanks | Number of tadpoles\newline 1 | 10\newline 2 | 20\newline 3 | 30\newline 4 | 40\newline 5 | ?  & Each tank has 10 tadpoles. How many tadpoles are in 5 tanks? & To solve this problem, we need to complete the table according to the tendency of the columns. \\
    \midrule
    7 &  | Blue sticker | Green sticker\newline Front door of the house | 2 | 4\newline Back door of the house | 3 | 3 & Lester keeps all his spare keys in a box under his bed. Recently, Lester decided the box was becoming unmanageable, as none of the keys were labeled. He set about labeling them with colored stickers that indicated what each key opened. What is the probability that a randomly selected key opens the front door of the house and is labeled with a green sticker? Simplify any fractions. & To solve this problem, we need to find the number of outcomes in the event and the total number of outcomes. Then compute the probability. \\
    \midrule
    8 & Sparrowtown | 8:00 A.M. | 2:00 P.M. | 4:45 P.M.\newline Danville | 9:15 A.M. | 3:15 P.M. | 6:00 P.M.\newline Princeton | 10:30 A.M. | 4:30 P.M. | 7:15 P.M.\newline Westminster | 11:45 A.M. | 5:45 P.M. | 8:30 P.M.\newline Oakdale | 1:30 P.M. | 7:30 P.M. | 10:15 P.M. & Look at the following schedule. Lee just missed the 4.30 P.M. train at Princeton. What time is the next train? & To solve this problem, we need to locate the entry from the table and read the next entry. \\
    \bottomrule
    \end{tabular}%
}
    \caption{Skill description annotation for TabMWP dataset.}
    \label{tab:skill_label_tab}
\end{table}
\begin{table}[t]
    \centering
\resizebox{\columnwidth}{!}{%
    \begin{tabular}{c p{15cm} p{5cm}}
        \toprule
        ID & Question & Skill Description \\
        \midrule
        1 & Angela slept 6.5 hours every night in December. She decided she should get more sleep and began sleeping 8.5 hours a night in January. How much more sleep did Angela get in January? & To solve this question, we need to do subtraction, inference the total number of days in a month, and do multiplication. \\
        \midrule
        2 & Edith is a receptionist at a local office and is organizing files into cabinets. She had 60 files and finished organizing half of them this morning. She has another 15 files to organize in the afternoon and the rest of the files are missing. How many files are missing? & To solve this question, we need to do division, addition, and subtraction. \\
        \midrule
        3 & Rosalina receives gifts from three people on her wedding day. How many gifts did she get if Emilio gave 11 gifts, Jorge gave 6 gifts, and Pedro gave 4 gifts? & To solve this question, we need to do addition. \\
        \midrule
        4 & A store puts out a product sample every Saturday. The last Saturday, the sample product came in boxes of 20. If they had to open 12 boxes, and they had five samples left over at the end of the day, how many customers tried a sample if the samples were limited to one per person? & To solve this question, we need to do multiplication and subtraction. \\
        \midrule
        5 & Billy is counting the rings in two trees. Weather fluctuations in this area mean that each tree's rings are in groups of two fat rings and four thin rings. If Billy counts 70 ring groups in the first tree and 40 ring groups in the second tree, how much older is the first tree? (Trees grow 1 ring per year.) & To solve this question, we need to do addition, subtraction, and multiplication. \\
        \midrule
        6 & A group of six friends planned to buy a car. The cost of the car is \$1700 and they plan to share the cost equally. They had a car wash to help raise funds, which would be taken out of the total cost. The remaining cost would be split between the six friends. At the car wash, they earn \$500. However, Brad decided not to join in the purchase of the car. How much more does each friend have to pay now that Brad isn't participating? & To solve this question, we need to do subtraction, division, and multiplication. \\
        \midrule
        7 & In Fifi's closet, she hangs all of her clothes on colored plastic hangers.  She has clothes hanging on 7 pink hangers, 4 green hangers, one less blue hanger than there are green hangers, and one less yellow hanger than there are blue hangers.  What is the total number of colored hangers in Fifi's closet? & To solve this question, we need to do subtraction and addition. \\
        \midrule
        8 & At the family reunion, everyone ate too much food and gained weight.  Orlando gained 5 pounds.  Jose gained two pounds more than twice what Orlando gained.  Fernando gained 3 pounds less than half of what Jose gained.  How much weight, in pounds, did the three family members gain at their reunion? & To solve this question, we need to do multiplication, addition, and subtraction. \\
        \bottomrule
    \end{tabular}%
}
    \caption{Skill description annotation for GSM8K dataset.}
    \label{tab:skill_label_gsm8k}
\end{table}

For \texttt{Skill-KNN-zero} with zero-shot generation of the skill description, the prompts used for the four datasets are shown in Table \ref{tab:zero-shot-prompt}.
\begin{table}[tb!]
    \centering
\resizebox{\columnwidth}{!}{%
    \begin{tabular}{c|c}
    \toprule
        Dataset & Prompt \\
    \midrule
        TabMWP & Describe the required skills to solve the following problems based on the data from the tables in one sentence \\
    \midrule
        GSM8K & Describe the required skills to solve the following questions in one sentence \\
    \midrule
        Spider & Describe the needed skills to solve the task on the database schema in one sentence. \\
    \midrule
        COGS & Describe the required skills to parse the following sentences in one sentence. \\
    \bottomrule
    \end{tabular}%
}
    \caption{Prompts for zero-shot skill generation.}
    \label{tab:zero-shot-prompt}
\end{table}

\section{Analysis and Ablation}
\label{sec:analysis}
This section provides in-depth analysis and explains the reasoning of the success of \method.
\paragraph{Why reasoning skill is a better guidance for demonstration selection?}\quad
\begin{figure}[t]
    \centering
    \includegraphics[width=0.9\columnwidth]{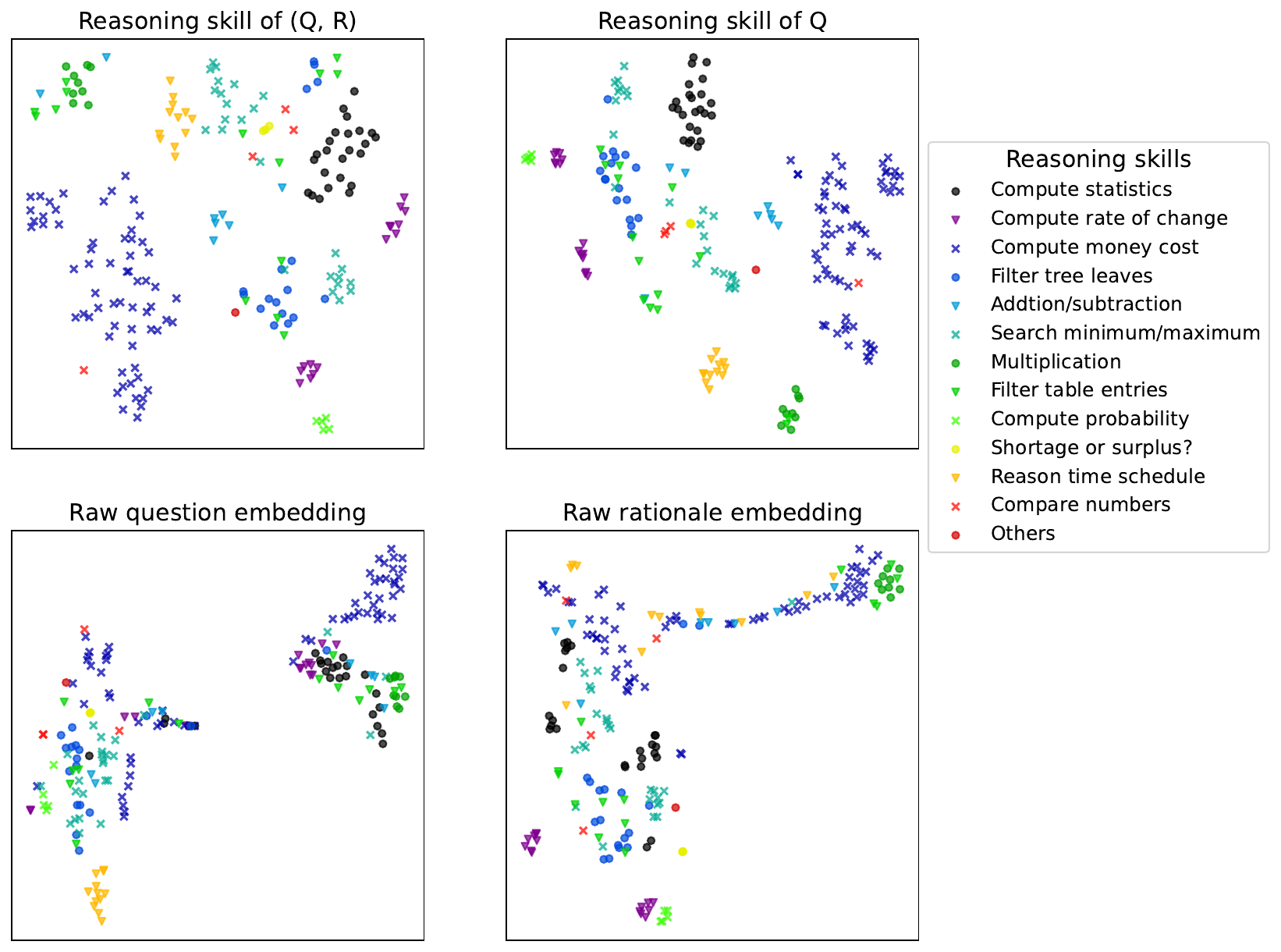}
    \caption{t-SNE projections of reasoning skills predicted from $(Q, \RA)$ (top-left), reasoning skills predicted from $Q$ (top-right), raw question embedding (bottom-left), and raw rationale embedding (bottom-right). The 12 different colors correspond to 12 skill labels provided by human.}
    \label{fig:clustering}
\end{figure}

In \textbf{TabMWP} dataset, 200 examples are labeled based on the skills being showcased out of 12 manually-crafted skills labels, including ``compute statistics'', ``compute rate of change'', ``Reason time schedule'', ``Compute probability'', et.~al. We investigate how the unsupervisedly discovered reasoning skills by \method align with human's understanding of skills. More specifically, a visualization of how human-labeled skills distribute based on the t-SNE projections of four different types of embedding is shown in Fig. \ref{fig:clustering}. Both the reasoning skill encoder (reasoning skill of $(Q, R)$) and the reasoning policy (reasoning skill of $Q$) trained by \method demonstrate clear separation of the labeled 12 skills. At the mean time, the human-labeled skills are not well-separated by raw question embedding, and even raw rationale embeddings. This indicates that the discovered reasoning skills aligns well with human-labeled skills even without explicit labels being provided during the training. This sheds the light on why the demonstration selection based on similar reasoning skills can improve the CoT prompting.

\begin{figure}[!h]
    \centering
    \begin{subfigure}{.48\textwidth}
        \includegraphics[width=\columnwidth]{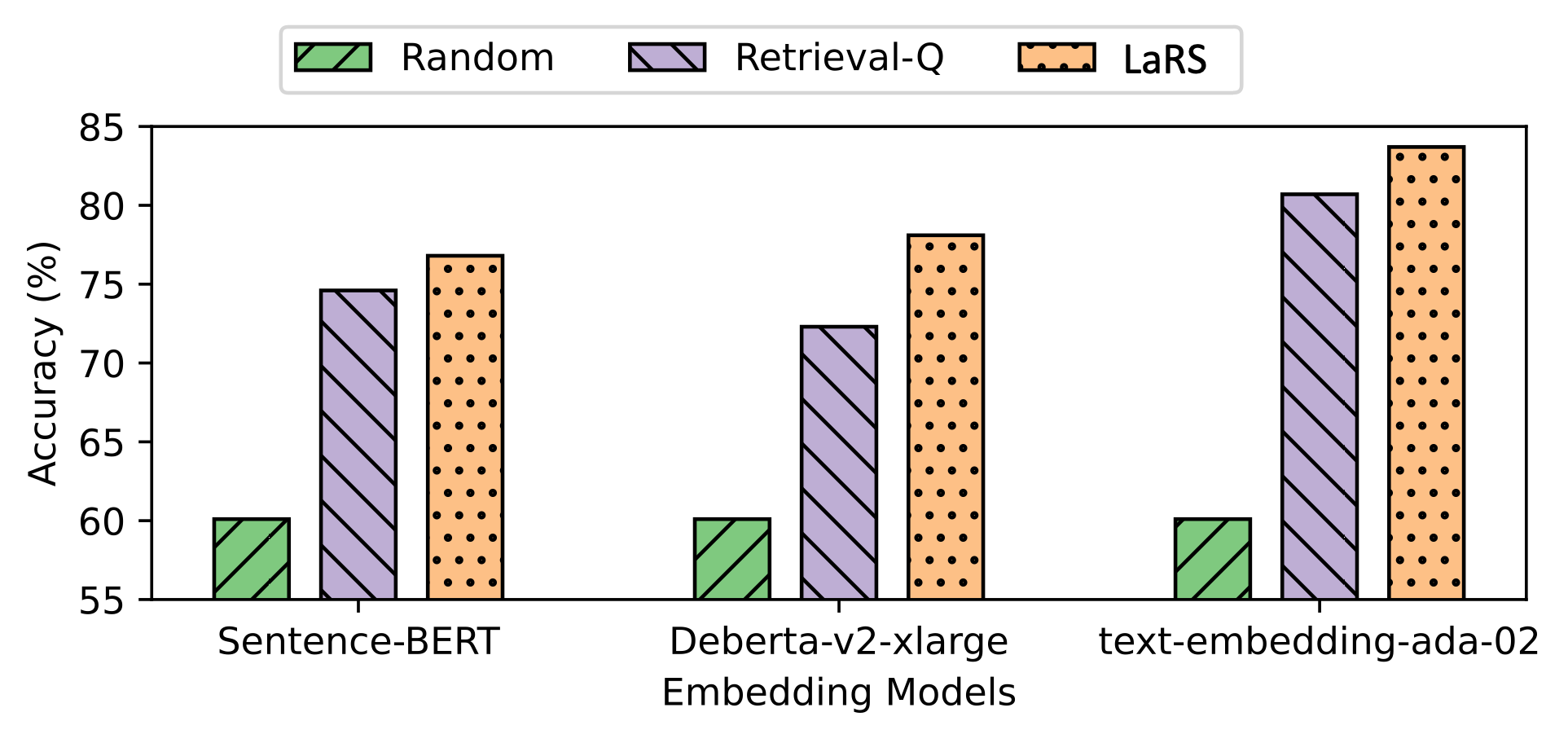}
        \caption{The accuracy of Random, \qmethod, and, \rmethod based on three different pre-trained embedding models.}
        \label{fig:embedding}
    \end{subfigure}\quad%
    \begin{subfigure}{.48\textwidth}
        \includegraphics[width=\columnwidth]{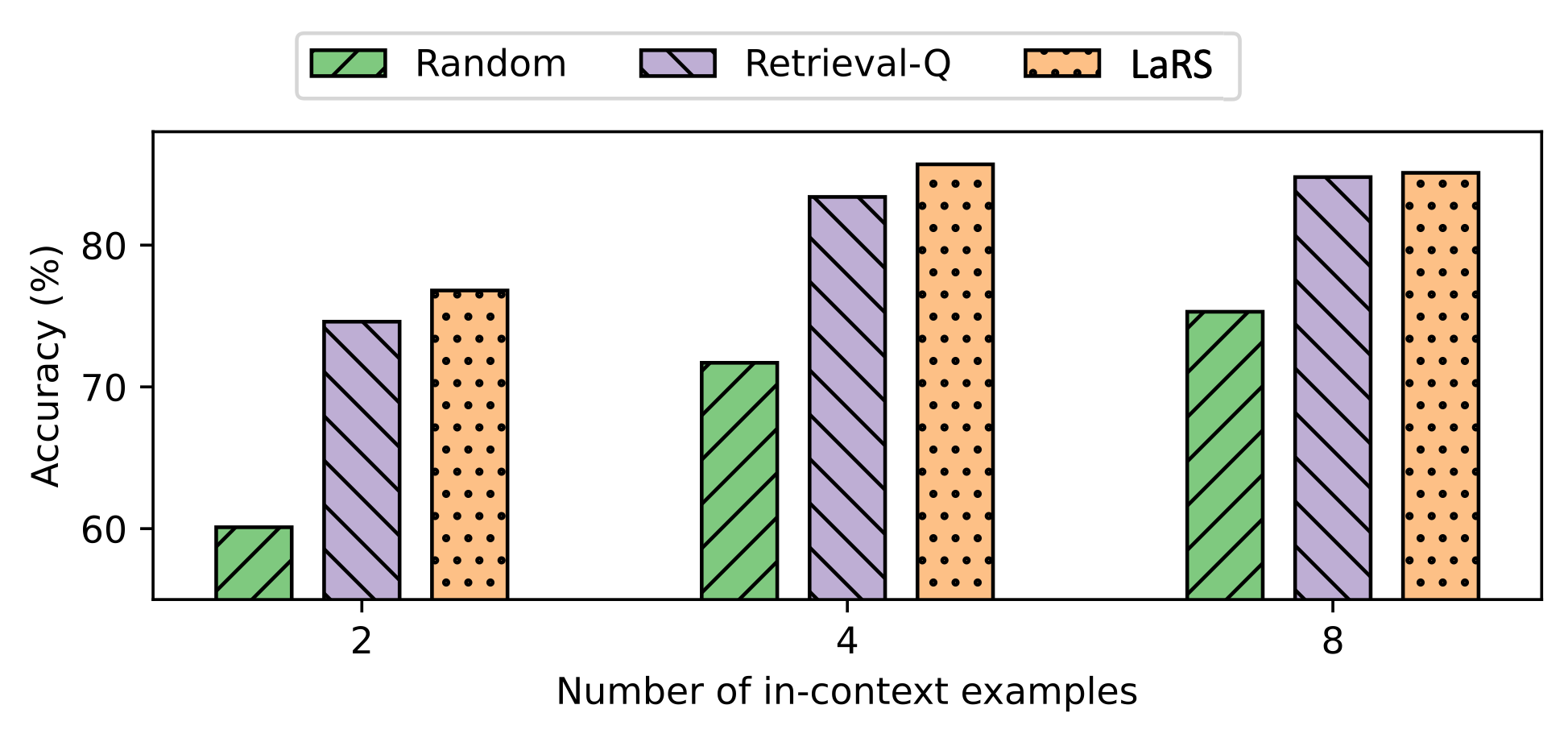}
        \caption{The accuracy of Random, \qmethod, and \rmethod  using different number of in-context examples.}
        \label{fig:num_exmaple}
    \end{subfigure}
    \caption{Performances of three different selection methods under (a) different pre-trained embedding models, and (b) different number of in-context examples.}
\end{figure}
\paragraph{Robustness to different pre-trained embedding models.}\quad Fig. \ref{fig:embedding} compares the performances of Random, \qmethod, and \rmethod based on three pre-trained embedding models, including Sentence-BERT~\cite{reimers2019sentence}, Deberta-v2-xlarge, and, text-embedding-ada-02~\cite{neelakantan2022text} from OpenAI. We observe that the performances of retrieval-based selection methods monotonously improve with more capable pre-trained embedding models. However, our \rmethod shows consistent improvements over \qmethod~given the same embedding models. 

\paragraph{Robustness to $k$: the number of in-context examples.} \quad This study compares three selection methods, including Random, \qmethod, and \rmethod under three different number of in-context examples 2, 4, and 8. The results are summarized in Fig. \ref{fig:num_exmaple}. While the accuracy monotonously improves with the increasing number of in-context examples, \rmethod consistently outperforms \qmethod.

\paragraph{How does Skill-KNN perform under stricter conditions?}
\begin{table}[t]
\centering
\resizebox{0.6\columnwidth}{!}{%
\begin{tabular}{lcccc}
\toprule
 Method                    & TabMWP                  & GSM8K                     & Spider                    & COGS                          \\
\midrule
             \multicolumn{5}{c}{Backbone: gpt-3.5-turbo }\\
\midrule 
  \texttt{Skill-KNN-large} & \textbf{78.3} \rise{15.9} & 75.0 \drop{0.7} & \textbf{58.4} \rise{11.6} & 94.6 \rise{27.2} \\  
  \texttt{Skill-KNN-small} & 75.5 \rise{13.2} & 74.9 \drop{0.8} & 37.3 \drop{9.5}~ & 79.9 \rise{12.7} \\
  \texttt{Skill-KNN-zero} & 77.7 \rise{15.3} & 75.0 \drop{0.7} & 49.0 \rise{2.2}~ & 77.9 \rise{10.8} \\
  
                                     \cellcolor{gray!20} \rmethod  (ours)   &  \cellcolor{gray!20}\textbf{78.1} \rise{15.7} & \cellcolor{gray!20} \textbf{76.8} \rise{1.1} & \cellcolor{gray!20}\textbf{53.0} \rise{6.2}~ &\cellcolor{gray!20} \textbf{94.8} \rise{27.2}    \\ 
\midrule
             \multicolumn{5}{c}{Backbone: gpt-4o }\\
\midrule 
  \texttt{Skill-KNN-large}& 80.6 \rise{11.3}& 62.0 \drop{0.2} & 56.3 \rise{9.8} & \textbf{96.8} \rise{23.4}\\
  \texttt{Skill-KNN-small} & 77.4 \rise{8.1}~ & 62.3 \rise{0.1} & 47.4 \rise{0.3} & 79.4 \rise{6.0}~ \\
  \texttt{Skill-KNN-zero} & ~87.7 \rise{0.1}~ & \textbf{78.6} \drop{0.5} & 76.6 \rise{2.5} & 78.1 \rise{5.1}~ \\
                                    \cellcolor{gray!20}\rmethod (ours)     &\cellcolor{gray!20} \textbf{87.9} \rise{0.3} &\cellcolor{gray!20} 78.3 \rise{0.2}~ & \cellcolor{gray!20} \textbf{77.2} \rise{3.1}~          & \cellcolor{gray!20}  \textbf{90.2} \rise{17.2}  \\ 

\midrule
             \multicolumn{5}{c}{Backbone: claude-3-sonnet }\\
\midrule 
  \texttt{Skill-KNN-large} & & 93.2 \drop{0.1} & \textbf{25.9} \rise{7.6} & 96.2 \rise{17.0} \\
  \texttt{Skill-KNN-small} &  &  92.3 \drop{1.0} & 18.2 \drop{0.1} & 86.6 \rise{7.4} \\
  \texttt{Skill-KNN-zero} & 93.1 \rise{0.5} & 92.1 \drop{1.2}~ & 61.9 \rise{0.2} & 86.6 \rise{7.4}  \\
                                \cellcolor{gray!20} \rmethod (ours)     & \cellcolor{gray!20} \textbf{93.7} \rise{1.1}~  &\cellcolor{gray!20} \textbf{93.6} \rise{0.3}~ & \cellcolor{gray!20} \textbf{62.2} \rise{0.5}~ &\cellcolor{gray!20} ~\textbf{96.9} \rise{17.7}     \\ 
\midrule
             \multicolumn{5}{c}{Backbone: claude-3-haiku}\\
\midrule 
\texttt{Skill-KNN-zero} & 93.3 \rise{4.7} & ~\textbf{88.8} \rise{0.2}~ & 61.0 \rise{0.8} & ~79.7 \rise{13.5}  \\
\cellcolor{gray!20} \rmethod (ours)     & \cellcolor{gray!20} \textbf{93.3} \rise{4.7}~  &\cellcolor{gray!20} 87.6 \drop{1.0}~ & \cellcolor{gray!20} \textbf{61.3} \rise{1.1}~ &\cellcolor{gray!20} ~\textbf{89.9} \rise{23.7}     \\ 
\midrule
             \multicolumn{5}{c}{Backbone: Falcon-40B-Instruct }\\
\midrule 
                                    \texttt{Skill-KNN-large}                 & 55.9 \rise{10.2}           & \textbf{40.3} \rise{1.5} & 23.7 \rise{2.9} & 81.0 \rise{35.9}\\
                                    \texttt{Skill-KNN-small}                 & 51.4 \rise{5.7}~ & 36.5 \drop{2.3} & 20.3 \drop{0.3}~ & 59.4 \rise{14.3}   \\
                                    \texttt{Skill-KNN-zero} & 55.2 \rise{9.5}~ & 38.7 \drop{0.1} & 23.3 \rise{2.7} & 82.1 \rise{37.0}  \\
                                    \cellcolor{gray!20} \rmethod (ours)     &\cellcolor{gray!20} \textbf{57.7} \rise{12.0}& \cellcolor{gray!20}\textbf{39.1} \rise{0.3} & \cellcolor{gray!20}\textbf{24.8} \rise{4.2}~ &\cellcolor{gray!20} \textbf{89.5} \rise{44.4}    \\ 
\bottomrule
\end{tabular}%
}
\caption{\texttt{Skill-KNN-large}, \texttt{Skill-KNN-small}, and \texttt{Skill-KNN-zero} compare with \method.}
\label{tab:results}
\vspace{-8pt}
\end{table}

\section{Case Study}
To explore the examples categorized as distinct skills within the learned latent reasoning skill representation, we employed K-means clustering on the latent reasoning skills of 1,000 examples from the \textbf{TabMWP} dataset. The centroids of these clusters are detailed in Table \ref{tab:case_study1}. The analysis presented in this table reveals that our method effectively discerns examples showcasing specific skills, such as ``Searching minimum/maximum'' and ``Computing rate change''. 
\begin{table}[t]
\centering
\resizebox{\columnwidth}{!}{%
    \begin{tabular}{c p{6cm} p{10cm} c}
    \toprule
    Cluster ID & Table & Question & Skill \\
    \midrule
    0 & [TITLE]: School play committees\newline Committee | Boys | Girls\newline Casting | 17 | 5\newline Set design | 14 | 17\newline Lighting | 20 | 20\newline Costume | 7 | 4\newline Music | 2 | 13 & Some students at Dayton Middle School signed up to help out with the school play. Which committee has the most boys?\newline Options: (A) set design (B) lighting (C) casting (D) costume & Search minimum/maximum \\
    \midrule
    1 & [TITLE]: Pairs of shoes per store\newline Stem | Leaf \newline 1 | 9\newline 2 | 3, 3\newline 3 | 0, 2\newline 4 | 2, 4\newline 5 | 5, 7\newline 6 | 2, 5\newline 7 | 7\newline 8 | 0, 2, 4, 4\newline 9 | 0, 0 & Ivan counted the number of pairs of shoes for sale at each of the shoe stores in the mall. How many stores have exactly 23 pairs of shoes? & Search tree leaves \\
    \midrule
    2 & [TITLE]: None\newline piece of licorice | \$0.07\newline gum drop | \$0.05\newline gumball | \$0.08\newline cinnamon candy | \$0.01\newline peppermint candy | \$0.08\newline lemon drop | \$0.07 & Derek has \$0.06. Does he have enough to buy a piece of licorice and a cinnamon candy?\newline Options: (A) yes (B) no & Compute money cost \\
    \midrule
    3 & [TITLE]: None\newline Number of offices | Number of chairs\newline 1 | 2\newline 2 | 4\newline 3 | 6\newline 4 | 8\newline 5 | ? & Each office has 2 chairs. How many chairs are in 5 offices? & Multiplication \\
    \midrule
    4 & [TITLE]: None\newline popcorn balls | \$1/kilogram\newline coffee cake | \$3/kilogram\newline blueberry bars | \$2/kilogram\newline cream cheese bars | \$2/kilogram\newline lemon bars | \$3/kilogram & Sarah went to the store and bought 2 kilograms of blueberry bars. How much did she spend? (Unit: \$) & Compute money cost \\
    \midrule
    5 & [TITLE]: None\newline x | y\newline 12 | 19\newline 13 | 9\newline 14 | 2 & The table shows a function. Is the function linear or nonlinear?\newline Options: (A) linear (B) nonlinear & Compute rate of change \\
    \midrule
    6 & [TITLE]: Tractors\newline Farmer | Number of tractors\newline Farmer Judy | 4\newline Farmer Joe | 7\newline Farmer Megan | 7\newline Farmer Rick | 4\newline Farmer Jane | 4 & Some farmers compared how many tractors they own. What is the mode of the numbers? & Compute statistics \\
    \midrule
    7 & [TITLE]: None\newline pink sweater | \$6.69\newline pair of brown pants | \$9.66\newline plaid scarf | \$2.45\newline pair of sandals | \$7.69\newline white polo shirt | \$4.86 & How much money does Heather need to buy a pair of brown pants and a plaid scarf? (Unit: \$) & Compute money cost \\
    \midrule
    8 & [TITLE]: Tour bus schedule\newline Location | Arrive | Depart\newline the riverfront | 9:55 A.M. | 10:20 A.M.\newline the zoo | 10:35 A.M. | 11:30 A.M.\newline art museum | 12:05 P.M. | 12:30 P.M.\newline science museum | 1:00 P.M. | 1:45 P.M.\newline skyscraper | 1:50 P.M. | 2:20 P.M.\newline governor's mansion | 2:50 P.M. | 3:45 P.M.\newline old building | 4:00 P.M. | 4:45 P.M.\newline famous bridge | 5:15 P.M. | 5:40 P.M.\newline the aquarium | 6:20 P.M. | 7:00 P.M.\newline landmark sculpture | 7:45 P.M. | 8:20 P.M. & Look at the following schedule. Which stop does the bus depart from at 11.30 A.M.?\newline Options: (A) zoo (B) riverfront (C) old building (D) science museum & Reason time schedule \\
    \midrule
    \end{tabular}%
}
\end{table}
\begin{table}[t]
\centering
\resizebox{\columnwidth}{!}{%
    \begin{tabular}{c p{6cm} p{10cm} c}
    \toprule
    Cluster ID & Table & Question & Skill \\
    \midrule
    9 & [TITLE]: None\newline poppyseed muffin | \$2.31\newline bowl of yogurt | \$1.35\newline blueberry pancakes | \$7.28\newline hash browns | \$4.56\newline bowl of granola | \$2.97\newline bagel with cream cheese | \$2.56 & Max has \$13.33. How much money will Max have left if he buys a bagel with cream cheese and blueberry pancakes? (Unit: \$) & Compute money cost \\
    \midrule
    10 & [TITLE]: Balloons sold\newline Day | Number of balloons\newline Wednesday | 568\newline Thursday | 586\newline Friday | 558\newline Saturday | 565 & The manager of a party supply store researched how many balloons it sold in the past 4 days. On which day did the store sell the most balloons?\newline Options: (A) Wednesday (B) Thursday (C) Friday (D) Saturday & Search minimum/maximum \\
    \midrule
    11 & [TITLE]: None\newline forklift | \$9,987.00\newline dump truck | \$9,543.00\newline race car | \$8,370.00\newline crane | \$6,996.00\newline bulldozer | \$7,547.00\newline hydrofoil | \$8,047.00 & How much more does a forklift cost than a dump truck? (Unit: \$) & Compute money cost \\
    \bottomrule
    \end{tabular}%
}
    \caption{The closest examples to the 12 cluster centers computed by K-Means clustering method on reasoning skill latent variables.}
    \label{tab:case_study1}
\end{table}

\section{Theoretical Analysis}
\label{appendix:theory}
To prove Theorem~\ref{theorem:1}, we start with the equation of rationale generation via CoT prompting, employing the skill-based demonstration selection method denoted as $g_{skill}$. The process can be formalized as follows:
\begin{align}
    P_M(&\RA \mid Q, g_{skill}) = \int_{\mathcal{X}^k}P_M(\RA\mid pt)\Pi_{i=1}^k[g_{skill}(Q_i, R_i\mid Q) d(Q_i, R_i)]
    \label{eq:1}
\end{align}
where Equation~\ref{eq:1} is integrated by substituting $pt = (Q_1, \RA_1, \cdots, Q_k, \RA_k, Q)$ as outlined in Equation~\ref{eq:cot_promting}, leading to:
\begin{align}
    P_M(&\RA \mid Q, g_{skill}) = \int_{\mathcal{Z}}P_M(\RA \mid z, Q) P_M(z \mid Q) \Pi_{i=1}^k[P_{skill}(z \mid Q)] dz
    \label{eq:2}
\end{align}
In this context, $P_{skill}(z \mid Q)$ is defined as:
\begin{align}
    P_{skill}(z \mid Q) = \int_{(Q', \RA') \in \mathcal{X}} P_M(z \mid Q', R') g_{skill}(Q', R'\mid Q) d(Q', R') dz'
    \label{eq:p_skill}
\end{align}
Substituting the Definition~\ref{def:retrieval_RSD} into Equation~\ref{eq:p_skill}, leading to:
\begin{align}
    P_{skill}(z \mid Q) = \int_{(Q', \RA') \in \mathcal{X}} \int_{z'\in \mathcal{Z}} P_M(z \mid Q', \RA') P_E(Q', R' \mid z') P_E(z' \mid Q) dz'
\end{align}
Applying Assumption~\ref{assump:1} into the above equation, replacing $P_M(z \mid Q', R')$ with $P_E(z \mid Q', R')$:
\begin{align}
    P_{skill}(z \mid Q) & = \int_{(Q', \RA') \in \mathcal{X}} \int_{z'\in \mathcal{Z}} P_E(z \mid Q', \RA') P_E(Q', R' \mid z') P_E(z' \mid Q) dz' \nonumber \\
                        & = \int_{z'\in \mathcal{Z}} \delta(z=z') P_E(z' \mid Q) dz' \nonumber \\
                        & = P_E(z \mid Q) 
    \label{eq:3}
\end{align}
By reintegrating the derived expression for $P_{skill}(z \mid Q)$ back into Equation~\ref{eq:2}, we arrive at:
\begin{align}
    P_M(\RA \mid Q, g_{skill}) = \int_{\mathcal{Z}}P_M(\RA \mid z, Q) P_M(z \mid Q) \Pi_{i=1}^k[P_E(z \mid Q)]dz
\end{align}
Take the limit of $k \rightarrow \infty$, above equation siplifies to:
\begin{align}
    P_M(\RA \mid Q, g_{skill}) = \int_{\mathcal{Z}}P_M(\RA \mid z, Q) P_E(z \mid Q) dz
\end{align}
Applying Assumption~\ref{assump:1} into the above equation, replacing $P_M(R \mid z, Q) $ with $ P_E(R \mid z, Q)$:
\begin{align}
    P_M(\RA \mid Q, g_{skill}) = \int_{\mathcal{Z}}P_E(\RA \mid z, Q) P_E(z \mid Q) dz = P_E(R \mid Q)
\end{align}
According to Assumption~\ref{assump:2}, the example bank can approximate expert rationale generation, or $P_E(R \mid Q) = P^*(R \mid Q)$, we then conclude:
\begin{align}
    P_M(\RA \mid Q, g_{skill}) = P^*(R \mid Q)
    \label{eq:proof_theorem1}
\end{align}
Equation~\ref{eq:proof_theorem1} means that the CoT prompting under the skill-based demonstration selection method give the optimal conditional distribution of rationales given questions by Definition~\ref{def:optimal}. This proves the Theorem~\ref{theorem:1} under Assumption~\ref{assump:2} and Assumption~\ref{assump:1}.

\end{document}